\documentclass[journal]{IEEEtran}

\usepackage{times}
\usepackage{epsfig}
\usepackage{graphicx}
\usepackage{amsmath}
\usepackage{amssymb}

\ifCLASSINFOpdf
\else
\fi
\begin{document}
%
\title{Deep Architectures and Ensembles for Semantic Video Classification}
%
%
%


\author{Eng-Jon Ong, 
Sameed Husain, 
Mikel Bober-Irizar, 
Miroslaw Bober$^{*}$
\thanks{Mikel Bober-Irizar was with Visual Atoms Ltd, Guildford, Surrey, UK.
E. Ong, S. Husain and M. Bober were with University of Surrey, Guildford, Surrey, UK.
E-mails: mikel@mxbi.net,\{e.ong,sameed.husain,m.bober\}@surrey.ac.uk }
}
%
%

\markboth{Journal of \LaTeX\ Class Files,~Vol.~14, No.~8, August~2018}%
{Shell \MakeLowercase{\textit{et al.}}: Bare Demo of IEEEtran.cls for IEEE Journals}
%



\maketitle

\begin{abstract}
This work addresses the problem of accurate semantic labelling of short videos. To this end, a multitude of different deep nets, ranging from traditional recurrent neural networks (LSTM, GRU), temporal agnostic networks (FV,VLAD,BoW), fully connected neural networks mid-stage AV fusion and others.
Additionally, we also propose a residual architecture-based DNN for video classification, with state-of-the art classification performance at significantly reduced complexity. 
Furthermore, we propose four new approaches to diversity-driven multi-net ensembling, one based on fast correlation measure and three incorporating a DNN-based combiner. We show that significant performance gains can be achieved by 
ensembling 
diverse nets and we investigate factors contributing to high diversity.  Based on the extensive YouTube8M dataset, we provide an in-depth evaluation and analysis of their behaviour. We show that the performance of the ensemble is state-of-the-art achieving the highest accuracy on the YouTube8M Kaggle test data. The performance of the ensemble of classifiers was also evaluated on the HMDB51 and UCF101 datasets, and show that the resulting method achieves comparable accuracy with state-of-the-art methods using similar input features.
\end{abstract}


%
\IEEEpeerreviewmaketitle


\section{Introduction}
Accurate clip-level video classification, utilising a rich vocabulary of sophisticated terms, remains a challenging problem.
One of the contributing factors is the complexity and ambiguity of the interrelations between linguistic terms and the actual audio-visual content of the video. For example, while a "travel" video can depict any location with any accompanying sound, it is the {\em intent of the producer} or even the {\em perception of the viewer} that makes it a "travel" video, as opposed to a "news" or "real estate" clip. Hence true {\em understanding} of the video's overall meaning is called for, and not mere {\em recognition} or a ``sum" of the constituent locations, objects or sounds. 

Another factor is the multi-dimensional (space and time) and multi-modal (audio and video) characteristics of the input data, which exponentially amplifies the complexity of the task compared to the already challenging problems of semantic annotation of images or audio snippets. For videos, a successful approach has to identify and localise important semantic entities not only in space, but also in time; it has to understand not only spatial but also temporal interactions between semantic entities or events, and it also has to link and balance the sometimes contradictory clues originating from the audio and video tracks. 

The recent Kaggle competition entitled "Google Cloud \& YouTube-8M Video Understanding Challenge" provided a unique platform to benchmark existing methods and to develop new approaches to video analysis and classification. The associated {\bf YouTube-8M (v.2)} dataset contains approximately 7 million individual video clips, corresponding to almost half a million hours (totalling 50 years!), annotated with a rich vocabulary of 4716 semantic labels \cite{45619}. The challenge is to develop classification algorithms which accurately assign video-level semantic labels. 

Given the complexity of the task, where humans are known to use diverse clues, we hypothesise that a successful solution must efficiently combine different expert models. Here, we pose several important questions: 
\begin{itemize}
\item What are the best architectures for this task? 
\item How to construct diverse models and optimally to combine them?
\item Do we need to individually train and combine discrete models or can we simply train a very large/flexible Deep Neural Networks (DNNs) to obtain a fully trained end-to-end solution? 
\end{itemize}

The first question clearly links to ensemble-based classifiers, where a significant body of prior work demonstrates that diversity is important. However, do we know all the different ways to promote diversity in DNN architectures? On the second question, our analysis shows that training a single network results in sub-optimal solutions as compared to an ensemble.

This paper is based on our work on the above Kaggle competition \cite{cultivate}. However, this work has significantly newer contributions and advances the field in a number of ways.
Firstly, we propose a deep residual architecture for semantic classification and demonstrate that it achieves state-of-the art classification performance  with significantly faster training and reduced complexity. 
Secondly, in order to advance beyond the state-of-the art, we propose four new approaches to ensembling of multiple classifiers.  We show a very simple but effective method which is based on optimal weight approximation and determined by a fast correlation measure. Further, we also propose and investigate three (learning) approaches incorporating DNN-based ensemblers. Our extensive experiments demonstrate that significant performance gains can be achieved by optimal ensembling of diverse nets and we investigate, for the first time, factors contributing to productive diversity. Based on the extensive YouTube8M dataset, we study and comparatively evaluate a broad range of deep architectures, including designs based on recurrent networks (RNN, LSTM), feature space aggregation (FV, VLAD, BoW), simple statistical aggregation, mid-stage AV fusion and others.  Finally, we show that our diversity-guided solution delivers a GAP of 85.12\% (on Kaggle evaluation set), which is the best result published to date. Importantly, our solution has a significantly reduced complexity compared to the previous state of the art.

\subsection{Related Work}
Recently, there has been significant progress in image recognition and detection using DNNs \cite{dnn_survey,met_det,dup_met}. However, additional temporal information and multiple frames exist when dealing with videos, resulting in different approaches to video classification.
We next review existing approaches to video classification before discussing ensemble-based classifiers. Initial approaches have relied on hand crafted spatio-temporal features, for example Dense Trajectories (DT) \cite{densetraj} and its improvement iDT \cite{iDT}. Ng et al. \cite{7299101} introduced two methods which aggregate frame-level features into video-level predictions: Long short-term memory (LSTM) and feature pooling. Fernando et al. \cite{7458903} proposed a novel rank-based pooling method that captures the latent structure of video sequence data. More recently, Xu et al. \cite{seqvlad} proposed SeqVLAD, where the temporal agnostic aggregation method of Vector of Locally Aggregated Descriptors (VLAD) \cite{7937898} was integrated into Gated Recurrent Units (GRUs). Multiple forms of spatio-temporal information ranging from dense trajectories, RGB-frame and optical flow information can be combined together with VLAD for video classification, as proposed by \cite{convpool}.

Karpathy et al. \cite{6909619} investigated several methods for fusing information across temporal domain and introduced Multiresolution CNNs for efficient video classification. 
Wu et al. \cite{Wu} developed a multi-stream architecture to model short-term motion, spatial and audio information respectively. LSTMs are then used to capture long-term temporal dynamics. 

DNNs are known to provide a significant improvement in performance over traditional classifiers across a wide range of datasets. However, it has also been found that further significant gains can be achieved by constructing ensembles of DNNs. One example is the ImageNet Large Scale Visual Recognition Challenge (ILSVRC) \cite{ILSVRC15}. Here, improvements up to 5\% were achieved over individual DNN performance (e.g. GoogLeNet\cite{googlenet}) by using ensembles of existing networks. Furthermore, all the top entries in this challenge employed ensembles of some form.

One of the key reasons for such a large improvement was found to be due to the diversity present across different base classifiers (i.e. different classifiers specialise to different data or label subsets)\cite{Hansen,Krogh95}. An increase in diversity of classifiers of equal performance will usually increase the ensemble performance. There are numerous methods for achieving this; such as random initialisation of the same models, or data modification using Bagging \cite{Breiman} or Boosting \cite{Boosting} processes. Recently, work was carried out on end-to-end training of an ensemble based on diversity-aware loss functions. Chen et al. \cite{NCL} proposed to use Negative Correlation Learning for promoting diversity in an ensemble of DNNs, where a penalty term based on the covariance of classifier outputs is added to the loss function. An alternative was proposed by Lee et al \cite{Mheads} based on the approach of Multiple Choice Learning (MCL) \cite{MCL}. Here, multiple DNNs are trained together based on a loss function that uses the final prediction chosen from the DNN with the lowest loss value.

\subsection{Contribution and Overview}
The rest of the paper is organised as follows: In Section \ref{sec:dnn_models}, we evaluate the performance of a wide range of different DNN architectures and video features. We also propose a novel DNN architecture that is inspired based on ResNet \cite{HeZRS16} that achieves state of the art performance for individual classifiers. We then provide detailed analysis on their individual performances across different classes to show that they are indeed diverse, and offer strong potential for ensembling (Section \ref{sec:class_dep_dnn_perf}). In order to advance state of the art, we propose four different methods for DNN ensembling leading to performance that is significantly higher than individual DNN classifiers (Section \ref{sec:ensmb}). We also provide an analysis of where improvements were obtained by an ensemble of classifiers. To compare against existing methods, we have also evaluated the performance of our ensemble using transfer learning on two popular datasets, UCF101 and HMDB51, in Section \ref{sec:transfer_exp}. 
Finally, we draw conclusions in Section \ref{sec:conclusions}.

\section{DNN Models}
\label{sec:dnn_models}
In this section, we describe the three different classes of DNN architectures that can be used for semantic labelling. The first is detailed in Section \ref{sec:fc_nn_model}, which consists of fully connected NNs that use feature vectors based on the mean and standard deviation of the frames in a video (i.e. the frames of each video are aggregated using mean and standard deviation operations). This approach has the advantage of working with a representation that is simpler in size and computational complexity. The next class of DNNs considered are the recurrent networks of LSTM and GRUs (Section \ref{sec:temp_model}). These have the advantage of being able to explicitly model the temporal nature of the data. Finally, we have a class of models that account for individual frames in a video via aggregation mechanisms that are agnostic to the temporal ordering in Section \ref{sec:temp_ag_model}.

\begin{figure*}[h!]
\centering
\includegraphics[width=0.9\textwidth]{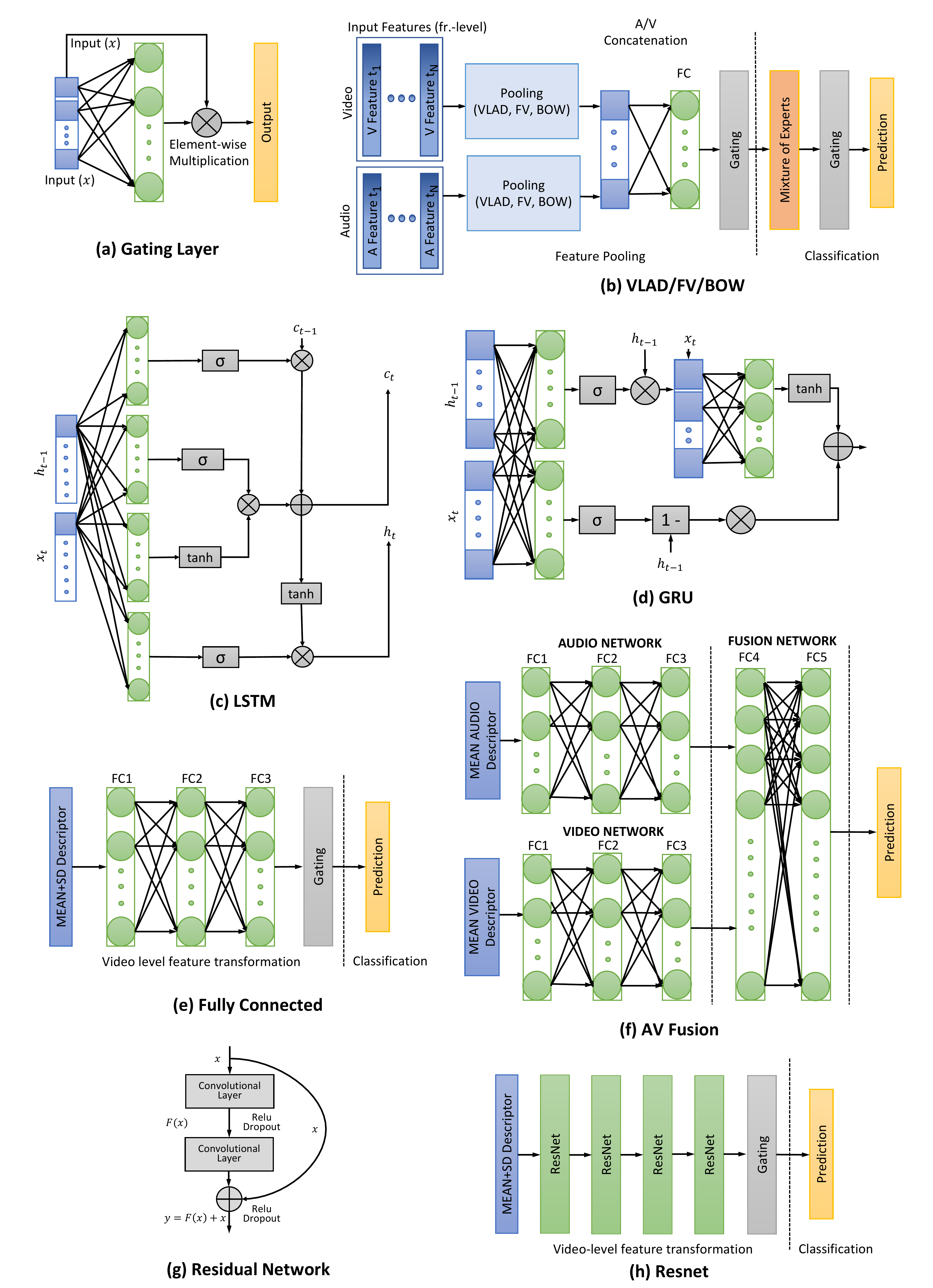}
\caption{This figure shows all the DNN components and architectures that are considered here (Section \ref{sec:dnn_models}).}
\label{fig:all-nets}
\end{figure*}

\begin{table*}
\caption{Summary of different architectures with their complexity, training time and the achieved GAP performance}
\centering
\begin{tabular}{|c|c|c|c|c|c|c|}
\hline
    Architecture & Architecture details & Params. &      Av. & Time   & Num. & Kaggle\\
                 &            & [$\times 10^6$] &  GAP  & (Hours) & Epochs & GAP\\
    \hline \hline
    ROI FC & FC Layers: 12K-12K-12K, DO: 0.3 & 502 &  81.84& 60 & 40 &81.74\\
    AV Fusion & (see Sec. \ref{sec:fcnnfeat}) & (see Sec. \ref{sec:fcnnfeat}) &  81.8 & 48 & 40 & 81.7\\ 
    \hline
    Gated ResNet 8K & 3 ResNet Layers: 8K-8K-8K-Gated, DO 0.4 &$238$ & 82.89 & 48 & 40 & 82.81\\
    Gated ResNet 10K & 2 (2-conv) ResNet Layers: 10K-10K-Gated, DO 0.3 &$238$ & 82.8 & 48 & 40 & 82.55\\
    Gated ResNet 10K & 2 (2-conv) ResNet Layers: 10K-10K-Gated, DO 0.4 &$238$ & 82.8 & 48 & 40 & 82.64\\
    \hline
    Gated FC 10K & 3 FC Layers: 10K-10K-10K-Gated, DO: 0.3 &$358$&  82.49 & 48 & 40 & 82.22 \\
    Gated FC 10K & 3 FC Layers: 10K-10K-10K-Gated, DO: 0.4 &$358$ &  82.77& 48 & 40 & 82.76 \\
    Gated FC 12K & 3 FC Layers: 12K-12K-12K-Gated, DO: 0.3 &$502$ & 82.41 & 60 & 40 & 82.22 \\
    Gated FC 12K & 3 FC Layers: 12K-12K-12K-Gated, DO: 0.4 &$502$ &  82.72 & 60 & 40 & 82.70 \\
    \hline
    LSTM & 2 layers, 1024 cells & 31 &  81.88 & 160 & 12 & 81.84 \\
    GRU & 2 layers, 1200 cells & 34 &82.15  & 199 & 9 & 82.07 \\
    \hline
    Gated BOW & 4096 Clusters & 39 & 81.97 & 348 & 20 & 81.79 \\
    Non-Gated BOW & 8000 Clusters & 17 &81.88 & 178 & 16 & 81.81\\
    Gated netVLAD & 256 Clusters & 308 & 82.89 & 304  & 12 & 82.83 \\
    Gated netRVLAD & 256 Clusters & 307 & 82.95 & 140 & 12 & 82.85 \\
    Gated netFV & 128 Clusters & 308 & 82.37 & 131 & 12  & 82.23\\
    \hline
\end{tabular}
\label{tab:dnn_perf}
\end{table*}
\subsection{Gated Fully Connected NN Architecture}
\label{sec:fc_nn_model}
For our work, we use a 3-hidden layer fully connected neural network, with layers FC6, FC7 and FC8. The size of the input layer is dependent on the feature vectors chosen. We also employ dropout on each hidden layer. These will be described in more detail in the sections below.

The activation function for all hidden units is ReLU. 
We have considered numbers of hidden units of 8000, 10000 and 12000 (also referred to as 8k, 10k and 12k), with droupout rates of 0.3 and 0.4.
The output layer is again a fully connected layer, with a total of 4716 output units, one for each class. In order to provide class-prediction probability values, each output unit uses a sigmoid activation function.

We have found significant improvements can be obtained when a simple fully connected output layer is replaced by a gating layer, as was found in \cite{MiechLS17}. 
An illustration of this layer can be seen in Fig. \ref{fig:all-nets}a. The structure is similar to a fully connected layer with two important features: 1) The number of hidden units is the same as the input dimension; 2) the input vector is multiplied element-wise with the hidden layer output, resulting in the output vector. Essentially, the hidden layer here acts as a ``gate'', determining how much of a particular input dimension to let through. Crucially, this gate is calculated using all the input values. This allows this layer to learn and exploit correlations (or de-correlations) amongst different classes.

\subsubsection{FC-NN Features}
\label{sec:fcnnfeat}
We employed the following types of input features:

\begin{itemize}
\item{\bf Video-Level Mean Features (MF)}\\
The frame-level features were obtained by two separate Inception DNNs, one for video and another for audio. They gave 1024 and 128 output values respectively, which are then concatenated into a 1152-dimensional frame level feature.
 The mean feature $\mu^I$ for each video was obtained by averaging these frame-level features across the time dimension. 

\item{\bf Video-Level Mean Features + Standard Deviation (MF+STD)}\\
We extract the standard deviation feature $\sigma^I$ from each video. The signature $\sigma^I$ is L2-normalised and concatenated with L2-normalised mean feature $\mu^I$ to form a 2304-dim representation $\phi^I=[\mu^I;\sigma^I]$. 

\item{\bf Region of Interest pooling (ROI)}\\
The ROI-pooling based descriptor, proposed by Tolias et al \cite{Radenovic2016CNNIR}, is a global image representation for image retrieval and classification. We compute a new video-level representation using the ROI-pooling approach, where the frame-level features are max-pooled across 10 temporal-scale overlapping regions, obtained from a rigid-grid covering the frame-level features, producing a single signature per region. These region-level signature are independently L2-normalised, PCA transformed and whitened. The transformed vectors are then sum-aggregated and L2-normalised. The dimensionality of final video-level representation is 1152, similar to that of the video-level mean features.

\item{\bf Audio-Visual Fusion network} (AVF)\\
The idea behind the AVF network is to perform audio-vision fusion in order to maximise information extracted from each mode. This method comprises of two stages: (i) First, the audio and visual features networks are trained separately to minimise the classification loss and then (ii) combined in a fusion network consisting of two fully-connected layers \cite{7956190}.

We first train the audio and video networks individually. We use three fully connected layers similar to FC6, FC7 and FC8, respectively, all of size 4096. Each FC layer is followed by a ReLU and a dropout layer. The output of the FC8 layer is passed through another fully connected layer FC9 which computes the predictions and finally updates the network parameters to minimise the cross entropy loss over the training data.

After training the audio and video networks, we discard their FC9 layers and connect their FC8 layers to the fusion network shown in Fig. \ref{fig:all-nets}(f). In this way, 4096-dim audio and 4096-dim video features are concatenated to form a 8192-dim representation as an input to the fusion network. This fusion network contains two fully connected layers of size 8192, followed by a fully connected prediction layer and cross-entropy optimisation.
%
\end{itemize}

\subsection{Temporal Models: LSTMs and GRUs}
\label{sec:temp_model}
In order to explicitly model the video frames as a temporal sequence, two recurrent neural networks were used: Long-Short-Term Memory (LSTM) \cite{Gers2000} and Gated Recurrent Units (GRU) \cite{cho14}. Both LSTMs and GRU encode information seen in the past using high dimensional vectors called memory cells. These memory cells have the same dimensionality as the LSTM and GRU output vectors.

In the LSTM model (Fig. \ref{fig:all-nets}c), the input and memory cell vectors are linearly transformed (via weight matrices and bias vectors), with a sigmoid function applied to yield three different vectors used for gating. These gates are applied by means of element-wise multiplication with another vector and are named the input, forget and output gates. The gates, together with the current memory state, input vector and previous output vector, are recursively used to update the memory state for each time step and also produce the current output vector.
In this paper, a stack of two LSTM layers were used, where the output of an initial LSTM is fed as input to a second LSTM unit, and the output of the second LSTM layer is used.
The output size (and memory state) of the LSTMs is 1024.


GRUs can be thought of as a simplified version of the LSTM, where the output gate is removed. Additionally, GRUs do not contain an internal hidden memory state as in LSTMs. Instead, the information from previous frames are encoded directly in the output vector.
As a result, the GRU architecture contains a smaller number of parameters compared to LSTMs. Despite this, the performance of GRUs are often better than LSTMs.
As with LSTMs, a stack of two GRUs is used. The size of the output state of the GRUs is 1200.

The outputs of both the LSTMs and GRUs were then passed into a gated fully connected layer that provides the 4717 class output values.

\subsection{Temporal-Agnostic Aggregation Models: NetVLAD, DeepFV and DeepBOW}
\label{sec:temp_ag_model}
Another class of models is where the aggregation of separate frames in a video is performed ignoring their temporal order. The DNNs of this nature considered in this paper contain the following methods: VLAD \cite{7937898}, Fisher Vectors\cite{PerronninD07} and Bag-of- Words \cite{bow}. These approaches are respectively denoted as netVLAD, netFV and netBOW. They were also used in \cite{willow} for the Youtube8M Kaggle challenge.

\subsubsection{Deep VLAD: netVLAD} 
NetVLAD \cite{7937898} is a CNN architecture that is trainable in a end-to-end manner directly for computer vision tasks such as image retrieval, place recognition and action recognition. The NetVLAD network typically consists of a standard CNN (eg. VGG \cite{Simonyan14c}, RESNET \cite{7780459}) followed by a Vector of Locally Aggregated Descriptors (VLAD) layer that aggregates the final convolutional features into a fixed dimensional signature, with its parameters trainable via back-propagation.

The VLAD block encodes the positions of convolutional descriptors in each voronoi region by computing their residuals with respect to the nearest visual words. A pre-computed code-book of $n$ cluster centres is first computed offline using K-means clustering. The descriptors are soft-assigned to each cluster centre and the residual vectors are accumulated to obtain cluster-level representations. The final VLAD representation is obtained by concatenating all aggregated vector for all clusters $n$. The VLAD block can be implemented effectively using standard CNN blocks (Convolution, Softmax and Sum-pooling).  

\subsubsection{Deep Fisher Vectors: netFV}
Another popular method for generating global descriptors for image matching is the Fisher Vector (FV) method, which  aggregates local image descriptors (e.g. SIFT \cite{Lowe04}) based on the Fisher Kernel framework. A Gaussian Mixture Model (GMM) is used to model the distribution of local image descriptors, and the global descriptor for a video is obtained by computing and concatenating the gradients of the log-likelihoods with respect to the model parameters. One advantage of the FV approach is its encoding of higher order statistics, resulting in a more discriminative representation and hence better performance \cite{PerronninD07}. Here, we have used a model that learns the FV model in an end-to-end manner. 

\subsubsection{Deep Bag-of-Words: netBOW}
The Deep Bag-of-Words encoding is another temporal-agnostic representation constructed from frame-level descriptors by grouping similar audio and visual features into clusters (known as visual- or audio-words). A video sequence is represented as a sparse histogram over the vocabulary. The model tested uses soft-assignment strategy which has been shown to deliver better performance in AV retrieval and classification applications.  

\section{Residual-DNNs for Learning Semantic Video Content: Fully Connected ResNet}
Inspired by the success of ResNet \cite{HeZRS16} for image recognition, we propose a Fully Connected ResNet (FCRN) architecture to tackle the problem of video classification. The FCRN architecture is substantially lower in complexity than ResNet in terms of the layers (101 ResNet vs 2-4 FCRN) and does not require images as input. This allows us to use train this DNN on a very large dataset such as Youtube-8M.

More precisely, let $x$ be the video level features (Mean+Standard deviation) extracted from a video. The ResNet block can be defined as:
\begin{equation}
y = F(x,\{C_{i}\})+x
\label{eq:rs1}
\end{equation}
where, $y$ and $C$ are the outputs and the weights of the convolutional layer respectively. The function $F(x,\{C_{i}\})$ represents the residual mapping to be learned. The ResNet block is demonstrated in Figure \ref{fig:all-nets}g, $F=C_{2} (D\otimes \phi(C_{1}x))$ in which $\phi$ denotes the ReLU and $D$ is the randomly sampled dropout mask. The operation $F+x$ is computed by a shortcut connection and element-wise addition.

In the FCRN architecture (Fig. \ref{fig:all-nets}h), the input (Mean+Standard deviation feature vector) is first fed to a Fully Connected layer. The output of FC layer is then passed through a series of ResNet blocks, before being forwarded into a gated fully connected layer that provides the 4717 class output values.

In Table \ref{FCRN_layers}, we perform experiments to compute the optimum depth of the 8K-FCRN architecture. The size of the Fully Connected layers and the dropout value is fixed to 8K and 0.4 respectively. It can be observed that optimum performance is achieved with 3 ResNet blocks. 

\begin{table}[h]
	\caption{Performance of 8K-FCRN with different number of ResNet blocks}
	\label{FCRN_layers}
	\centering
	\begin{tabular}{|c||c|}
		\hline
	Number of ResNet blocks&  GAP  \  \\ \hline \hline
	8K-8K & 82.27   \\ \hline
	8K-8K-8K  & 82.89  \\ \hline
	8K-8K-8K-8K & 82.32   \\ \hline 
	\end{tabular}
\end{table}

We also performed experiments to find the optimum depth of 10K-FCRN network (DO=0.4). It can be observed from Table \ref{FCRN_layers_10k} that the 2 block 10K-FCRN outperforms both the 1 block and 3 block network.

\begin{table}[h]
	\caption{Performance of 10K-FCRN with different number of ResNet blocks}
	\label{FCRN_layers_10k}
	\centering
	\begin{tabular}{|c||c|}
		\hline
	Number of ResNet blocks&  GAP  \  \\ \hline \hline
	10K & 82.13   \\ \hline
	10K-10K & 82.80  \\ \hline
	10K-10K-10K & 82.49   \\ \hline 
	\end{tabular}
\end{table}

\section{Individual DNN Experimental Results}
The complete Youtube-8M dataset consists of approximately 7 million Youtube videos, each approximately 2-5 minutes in length, with at least 1000 views each. There are 4716 possible classes for each video, given in a multi-label form. For the Kaggle challenge, we were provided with 6.3 million labelled videos (i.e. each video was associated with a 4716 binary vector for labels). For test purposes, approximately 700K unlabelled videos were provided. The resulting class test predictions from our trained models were uploaded to the Kaggle website for evaluation. 

The evaluation measure used is called `GAP20'. This is essentially the mean average precision of the top-20 ranked predictions across all examples. To calculate its value, the top-20 predictions (and their corresponding ground-truth labels) are extracted for each test video. The sets of top-20 predictions for all videos are concatenated as well as the corresponding ground-truth labels into two global lists. Both lists are then sorted according to their prediction confidence values and mean average precision is calculated on the resulting list.

The performances of individual DNNs is summarised in Table. \ref{tab:dnn_perf}. 
Here, we show both the GAP20 scores on a validation set where ground-truth labels are available, as well as GAP20 scores on unseen test data, by uploading the test-data inferences to the Kaggle website. We observe that the performances of the different DNNs fall in the range of 82\% to 83\%. 

As expected, GRUs perform better than LSTMs.
We find the temporal agnostic models based on VLAD and Fisher Vectors consistently achieve high Kaggle GAP20 scores above 82\%. Interestingly, the BOW models achieve lower scores of under 82\%.

However, using considerably simpler mean and standard deviation features in the fully connected and ResNet models provide similar performance to the other, more complex models.
We find that using both mean and standard deviation features together gives the best results. We have found that only using the standard deviation features degrades the performance to low 40\% GAP20. Using the only the mean features without standard deviation results in a decrease of approximately 2-3\% in GAP20 accuracy.

We additionally performed experiments to find the optimum depth and size of the fully connected layers in the ResNet blocks. Table \ref{RS_table} demonstrates the impact of different ResNet architectures on the classification performance of our FCRN network. 
It can be seen that 4 residual blocks of 8K hidden units achieves the best performance of 82.89\%.

\begin{table}[h]
	\caption{Impact of different ResNet architectures on the classification performance of FCRN network}
	\label{RS_table}
	\centering
	\begin{tabular}{|c|c|c|}
		\hline
		FC layer size& Number of ResNet blocks& GAP (\%) \\ \hline \hline
		4096 $\times$ 4096     &	4& 		82.35 \\ \hline
		8192 $\times$ 8192    &	3&		82.89 \\ \hline
		10240 $\times$ 10240  &	2&		82.80 \\ \hline
	\end{tabular}
\end{table}

However, none of the individual DNN performances exceed 83\%. Nonetheless, we find that whilst the performances of the DNNs are comparable, different types of DNNs perform well (and conversely) on {\em different} sets of classes. This in turn will provide significant benefits in the GAP score after ensembling these individual DNNs together, combining the `strengths' of each classifier. To see this, we next provide an analysis on how well each DNN did across different classes. To achieve this, we next propose a measure for how accurate each DNNs relative to the final GAP20 score.

\subsection{Class Dependent DNN Performance Measure}
\label{sec:class_dep_dnn_perf}
In this section, we analyse the performance of individual DNNs in order to understand how they can contribute to improvements in the final ensembled system.
%
%
%
To achieve this, we calculate separate accuracy scores for each classifier on each video label. Whilst this is not exactly the GAP20 score, it is highly related. 

The classifier accuracy is based on `oracle' outputs, that is, for some given class and example, an oracle will inform us with 1 if that example's class label was output correctly from some classifier, and 0 vice versa. Given that our classifiers output probability values, their output values need to be first binarised. We chose a threshold of 0.5 for this, so that any output value from a classifier with a value greater than or equal to 0.5 will be equated to 1, and 0 vice versa.

Now, let the number of videos be $N$, number of classifiers be $D$ and total classes be $C$. For one of these $D$ classifiers, we obtain an oracle output matrix specific to it by comparing its thresholded output with the groundtruth label. Specifically, suppose the $d^{th}$ classifier has the binarised output matrix $C_d \in \{0,1\}^{N \times C}$, with each of its element denoted as $c_{d,i,j}, i = 1,2,...,N, j = 1,2,...,C$. Suppose as before, the groundtruth labels for class $k$ and example $n$ is denoted as $y_{k,n}$, then the oracle matrix for this classifier is denoted as $O_d \in \{0,1\}^{N\times C}$, with each element $o_{d,i,j} = \mathbb{I}(c_{i,j} = y_{i,j})$.

\subsubsection{Class Accuracy of Base DNNs}
The accuracy of class $i$ of some DNN with index $d \in \{1,...,D\}$, can be directly obtained using the oracle matrix as follows:
\begin{equation}
A(i,d) = \frac{1}{N}\sum^N_{j = 1} o_{d, i, j }
\end{equation}
The performance of the classifiers for the most frequent 100 classes can be seen in Fig. \ref{fig:gap_cls}. It can be seen from Fig. \ref{fig:gap_cls}a that it is difficult to distinguish between the accuracy curves for the different methods, indicating that all the classifiers in the ensemble perform similarly. This correlates well with the overall GAP scores shown in Table. \ref{tab:dnn_perf} where all the individual DNNs used had GAP20 scores that were fairly similar. We find that the accuracy for all classes are very high. This is due to the imbalance between the occurrence of a class (i.e. video label) and non-occurrence. That is, the majority of the videos considered will not have a specific video label associated with it. As an example, the most frequent label of ``Games'' only occurs in about 10\% of the videos.

\subsubsection{Mean Delta-Class Accuracy}
There are variations amongst individual DNN performances that can be exploited by ensembling for raising the final ensemble GAP score.
To see this more clearly, we will compare the values of $A(i,d)$ with the class mean accuracy curve: $\bar{A}(i) = \frac{1}{D}\sum_{d=1}^D A(i,d)$,  with $i \in \{1,...,C\}$.
The deviations of each classifier compared with the mean accuracy can then be obtained:
$\delta A(i,d) = A(i,d) - \bar{A}(i)$.

\begin{figure*}[t]
\centering
\begin{tabular}{cc}
\includegraphics[width=0.45\textwidth]{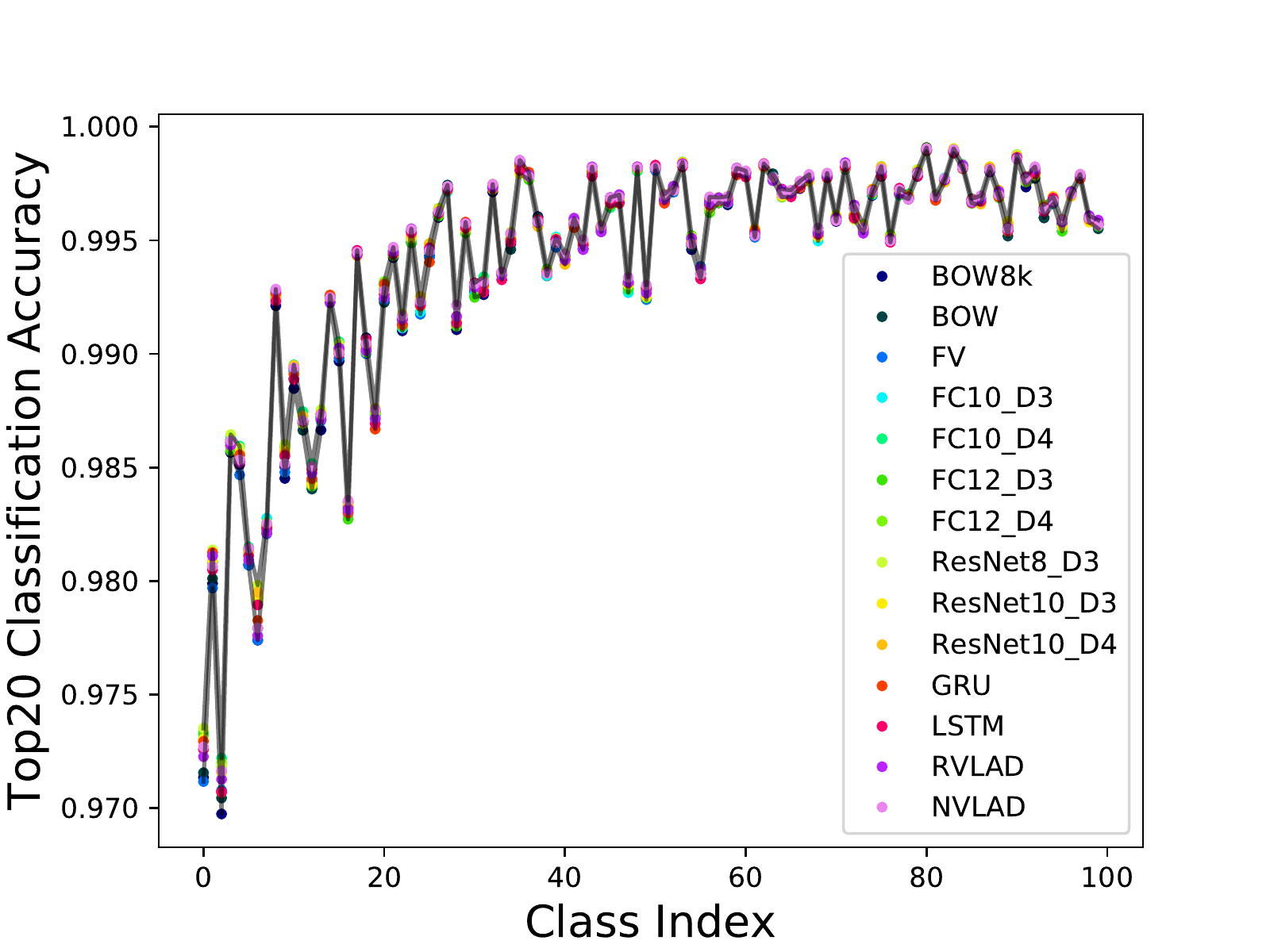} &
\includegraphics[width=0.55\textwidth]{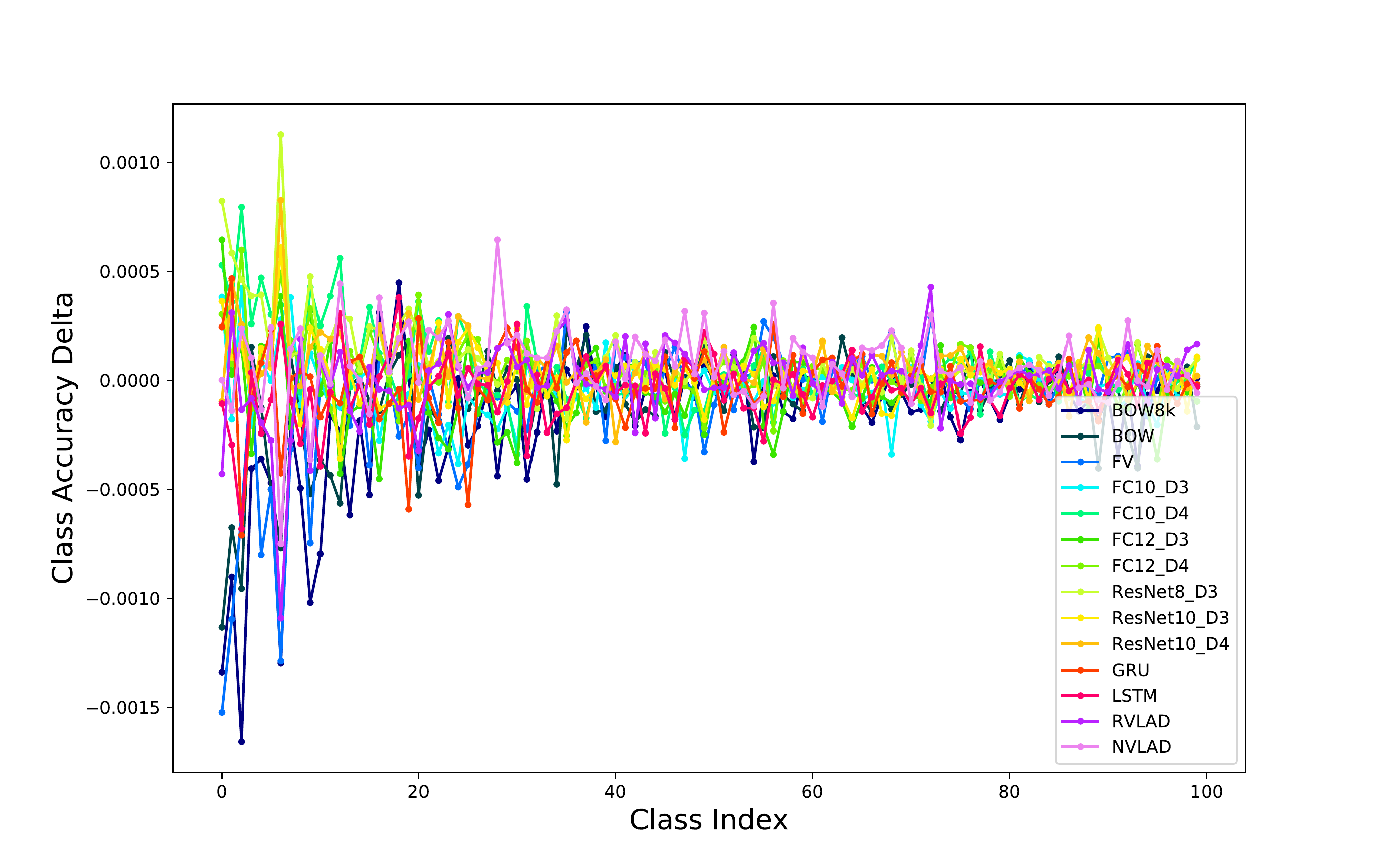}\\
(a)     & (b)
\end{tabular}
\caption{(a) Shows the top20 classification error for the first 100 classes. For clarity, we plot the maximum and minimum accuracy curves. This allows us to see the range of accuracies across different DNNs for a specific class. (b) shows the deviation of the errors of each DNN to the mean accuracy for the first 100 classes.}
\label{fig:gap_cls}
\end{figure*}

A plot of the deviation between the mean GAP accuracy and individual DNN performance is in Fig. \ref{fig:gap_cls}b. In this figure, we see the largest discrepancies between DNN performances exist at the most frequent labels, and decreasing with frequency. As a result, it becomes increasingly difficult to distinguish the classification accuracy delta curves in Fig. \ref{fig:gap_cls}b as the class frequency decreases.


In order to see the overall pattern, we extract a 
matrix from the per-class performance deviations $\delta A$. Each element $\delta a_{i,j}, i,j \in \{1,2...,D\}$ of this 
matrix is calculated as: 
\[
\delta a_{i,j} = \frac{1}{N} \sum^N_{k=1}\delta A(k,i)\delta A(k,j)
\]

\begin{figure}[t]
\centering
\includegraphics[width=0.5\textwidth]{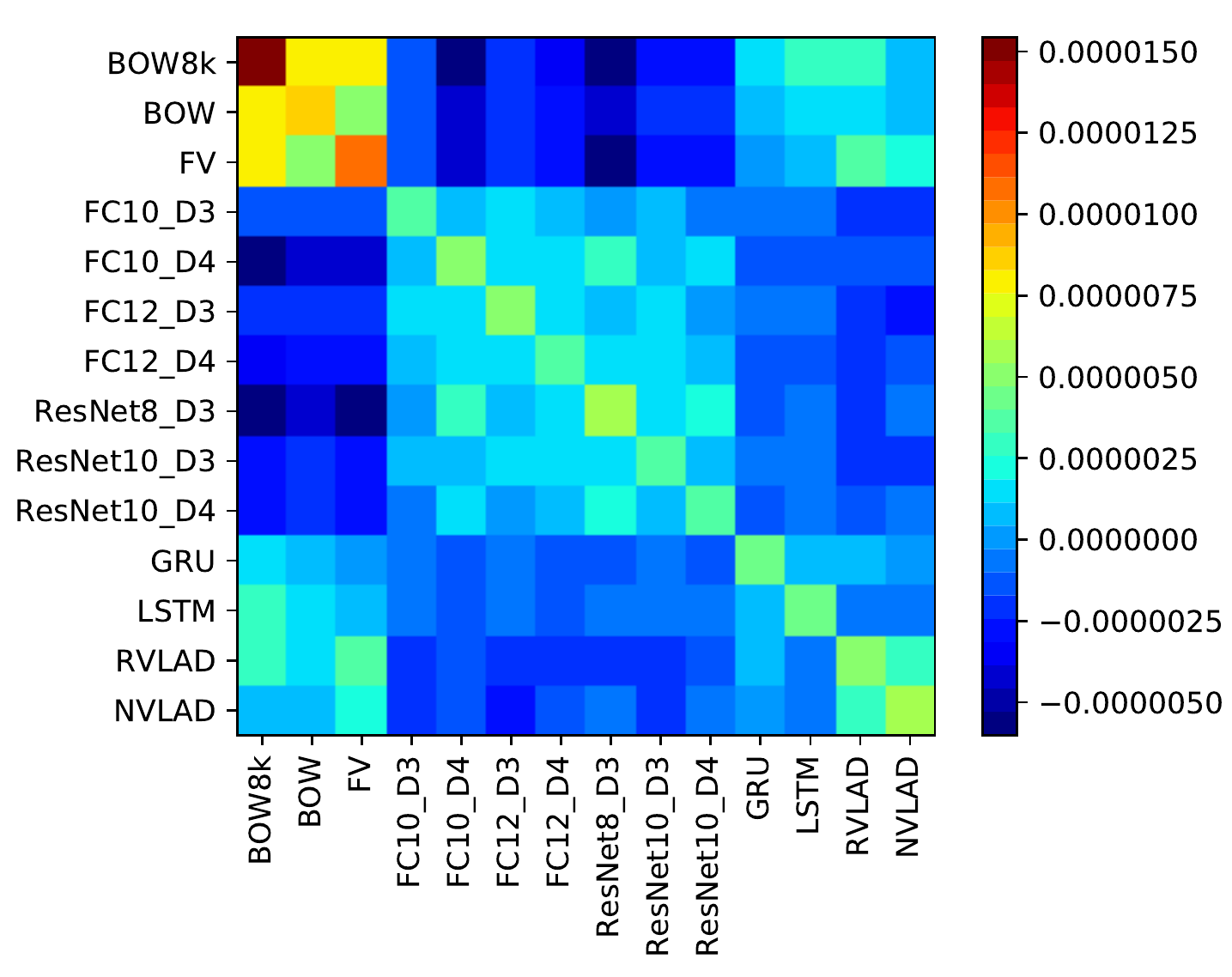}
\caption{This shows the matrix of the delta-per-class performance. The blocking in the matrix shows that DNNs of similar architecture perform similarly on similar classes.}
\label{fig:perclass_corr}
\end{figure}

The resulting 
matrix is shown in Fig. \ref{fig:perclass_corr}. Of particular interest are the negative deviations in the performances of pairs of individual DNNs (shown as dark blue). Negative values indicate that when a particular DNN is underperforming (below mean accuracy), whilst the other DNN over-performs. Here, we find that different classes of DNNs in the ensemble tend to contain negative correlations as can be seen in the dark blue entries in the correlation matrix. 

\begin{figure*}[h!]
\centering
\begin{tabular}{cc}
\includegraphics[width=0.5\textwidth]{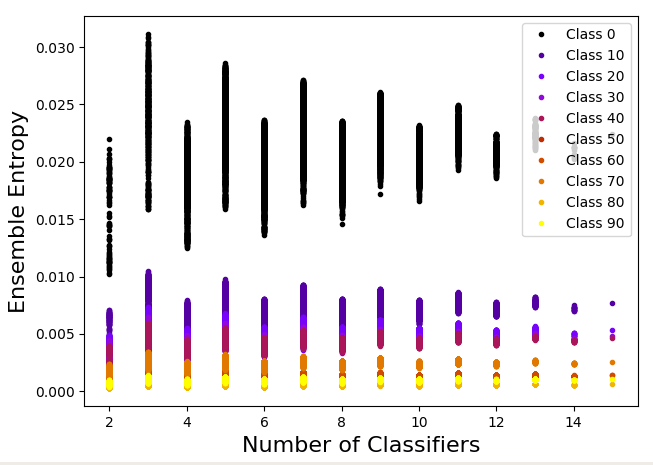} &
\includegraphics[width=0.5\textwidth]{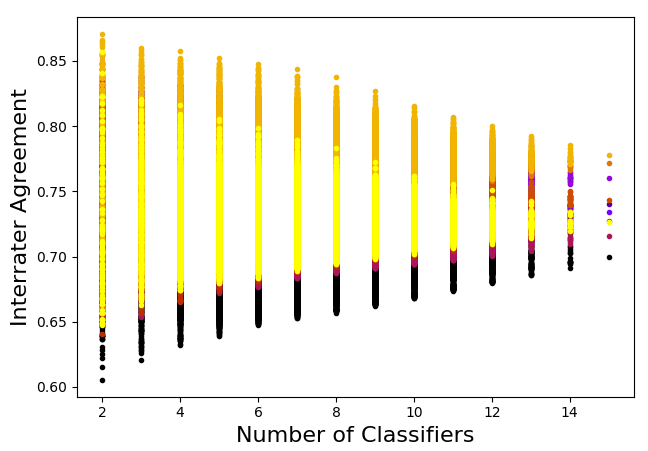}\\
(a)& (b)
\end{tabular}
\caption{ (a) Entropy diversity measure and (b) interrater agreement measures for different combinations of classifiers for the top 10 classes. The different colours in (a) and (b) represent different classes as indicated by the legend in (a).}
\label{fig:diversity}
\end{figure*}

\begin{figure*}[h!]
\centering
\includegraphics[width=0.9\textwidth]{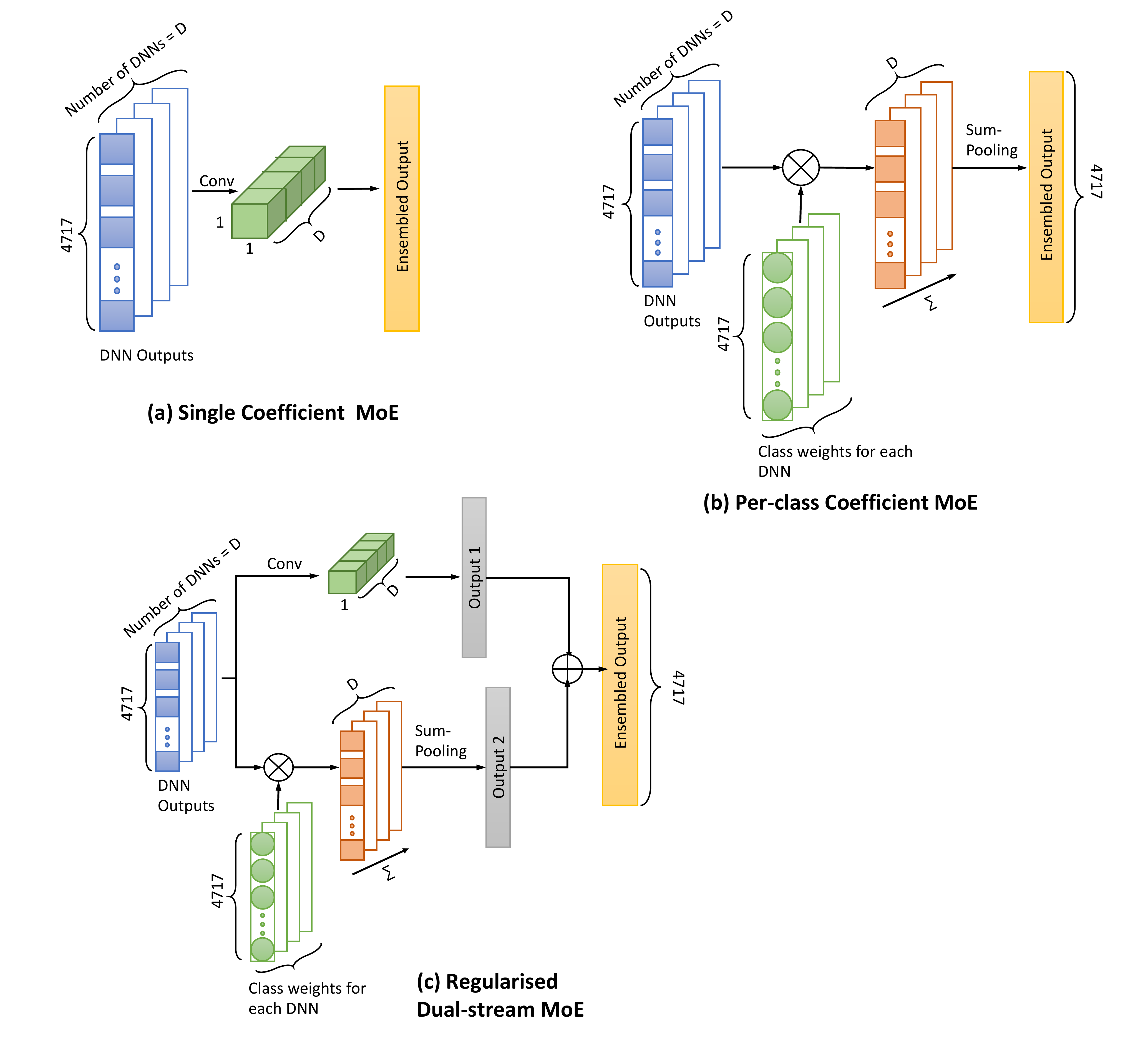}
\caption{Overview of 3 different ensembling architectures using (a) single coefficients per DNN, (b) separate coefficients per class and DNN and (c) regularised hybrid of (a) and (b) with two parallel streams. }
\label{fig:ensembling-nets}
\end{figure*}

\subsection{DNN Diversity Analysis}
It is well known that by combining the outputs from a set of base classifiers with similar performance, significant improvements in accuracy can be obtained. However, this improvement is governed by the diversity present in the group of the base classifiers. Roughly speaking, a set of classifiers are deemed diverse if they perform well on different examples or classes. However, there is no agreed diversity measure for a set of classifiers. 

Multiple authors have proposed a multitude of other potential diversity measures. Ten such measures were considered in \cite{Kuncheva2003}. It was observed there that there are correlations between the ensemble accuracy and amount of diversity present in the base-classifiers of an ensemble across the diversity measures analysed. These 10 measures can be split into two categories: pairwise and non-pairwise measures. Pairwise measures only quantify the diversity between two classifiers. To obtain a single number to represent an ensemble diversity from these measures, the pairwise average is usually obtained. Non-pairwise measures combine performances for all classifiers throughout the dataset into a single measure. The analysis here will use the non-pairwise measurement of Entropy and inter-rater agreement. 

One issue with the above measures is that they assume single label classification, Here, each video example is associated with multiple labels. Consequently, we have chosen to extract the diversity scores {\em per class}. The aim is to show how across different classes does the different diversity scores change as we add more base classifiers.

The diversity measures are all based on the oracle matrices described in Section \ref{sec:class_dep_dnn_perf}. They all use a common property, which is the number of classifiers that have recognised class $c$ in example $j$ correctly. We denote this as $l_{c,j} = \sum^N_{n=1}o_{d,j,c}$. We can also compute the average accuracy for class $c$ across all classifiers as:
\[
p_c = \frac{1}{DN} \sum^{D}_{i=1} \sum^{N}_{j=1} o_{i,j,c}
\]

We can then calculate the interrater agreement $\kappa$ for this ensemble of $N$ classifiers as:
\[
\kappa_c = 1 - \frac{\sum^{N}_{j=1} l_{c,j}(D-l_{c,j})}{ND(D-1)p_c(1-p_c)}
\]

The values of $\kappa_c$ indicate the amount of agreement between different classifiers in an ensemble, whilst correcting for chance. The smaller the value, the greater the diversity (since classifiers agree less with each other).

Another measure is based on entropy, and can be calculated as follows:
\[
E_c = \frac{\sum^N_{j=1}\min \{l_{c,j}, D - l_{c,j}\}}{N(D-\left \lceil{D/2}\right \rceil)}
\]
For this measure, a larger value will indicate larger diversity within the ensemble.
Using the above two measures, we exhaustively calculated the diversity measure for all possible combinations of the DNN classifiers in Table \ref{tab:dnn_perf} from a range of 2 classifiers to the final single ensemble of 14 DNNs. The results can be seen in Fig. \ref{fig:diversity}.

We find that in general, the lower bound for the results from the entropy measure increases. We have also shown the curve of average entropy across all possible combinations of DNNs for a fixed number of classifiers. This measure has an interesting artefact where odd numbered ensembles and even numbered ensembles have different ranges of entropy measurement, caused by the ceiling operator in the equation for $E_c$ above. We also observe that the upper bound for the interrater agreement decreases as the number of classifiers increase, indicating that the diversity of the ensemble is increasing. Interestingly, we observe that the less frequent the classes are, the less diverse the ensemble becomes.

\section{DNN Ensembling}
\label{sec:ensmb}
We have shown in Table \ref{tab:dnn_perf} that individual DNNs are not able to improve beyond 83\% GAP20 scores. Nonetheless, the analysis in Section \ref{sec:class_dep_dnn_perf} indicated that whilst the overall scores of individual DNNs are similar, the classes where they perform well in are quite diverse. This suggests that ensembling sets of individual DNNs will provide further improvements. 

A common approach to learn the ensembling coefficients is first to propose a diversity measure, such as those considered in \cite{Kuncheva2003} or \cite{Tang2006}. This is followed by simultaneously training multiple classifiers to also maximise the selected diversity measure. To this end, Lee et al. \cite{Mheads} proposed an end-to-end for training multiple classifiers simultaneously with the diversity measure factored into the loss function. This approach is unfeasible here, as the individual DNNs are complex and learn at different rates. Additionally, it is still an open question whether explicit optimisation of diversity scores will always lead to with increased ensemble accuracy. To analyse this, a theoretical study of 6 different diversity measures was carried out by Tang et al. \cite{Tang2006}. It was also found that diversity measures themselves can be ambiguous in predicting the generalisation accuracy of an ensemble.

Consequently, the approach chosen in this paper assumes that the base-classifiers in the ensemble are fixed (i.e. pre-learnt).
We have also chosen to perform ensembling by linearly combining outputs from each DNN. Therefore, the ensemble learning task is to determine the classifier coefficients.

The number of linear combination coefficients commonly fall into two classes: Single coefficient per-DNN and separate coefficients per (DNN, class) pair. Specifically, suppose the number of DNNs available is $N$. The outputs of each DNN for a $C$ class problem is denoted as $\mathbf{o}_i \in \mathbb{R}^C, i = 1,2,...,N$. The ensembling of the different approaches for the single coefficient per-DNN approach is then:
\begin{equation}
\mathbf{o}_{ens1} = \sum^{N}_{i=1} \alpha_i \mathbf{o}_i
\label{eq:ens1}
\end{equation}
where $\alpha_i$ are scalars. The second ensembling approach is:
\begin{equation}
\mathbf{o}_{ens2} = \sum^{N}_{i=1} \mathbb{\alpha}_i \circ \mathbf{o}_i
\label{eq:ens2}
\end{equation}
where $\circ$ represents element-wise multiplication and $\mathbf{\alpha}_i \in \mathbb{R}^C$ is the set of coefficients (one per class) for each DNN.
In order to learn the ensembling coefficients in Eq. \ref{eq:ens1} and Eq. \ref{eq:ens2}, two approaches are considered:
\begin{enumerate}
\item The first is a novel algorithm that attempts to optimise the ensembling coefficients by directly optimising the GAP20 score, as detailed in Section \ref{sec:correl_ensemble}.
\item The second approach poses the learning of ensembling coefficients as a Mixture of Experts (MoEs) problem, as detailed in Section \ref{sec:moe}.
\end{enumerate}

\subsection{Correlation-based Ensembling}
\label{sec:correl_ensemble}

In this section, we propose a novel non-DNN-based method for finding optimal DNN coefficients by greedy selection using Pearson's correlation of pairs of DNN outputs.
%
It attempts to directly optimise the GAP20 metric (or any arbitrary metric) unlike neural networks, which require a differentiable loss function such as cross-entropy to be used as a proxy for optimising GAP20.
\\\\
The algorithm works as follows:
\begin{enumerate}
  \item First, we compute a Pearson's correlation matrix based on the predictions of all the candidate DNN models.
  \item We greedily select the pair of DNN models A, B with lowest correlation as a starting point. We experimented with using other selection criteria and a greedy correlation-based selection gave the best results.
  \item For this pair of models, we compute the GAP score for $N$ combinations of weights (eg $w=\{0.2, 0.4, 0.6, 0.8\}$ for model A, and $1-w$ for model B). We used $N=4$ in our experiments.
  \item We can then fit a quadratic curve to the GAP score against weight for this pair of models.
  We found experimentally that across several evaluation metrics, the score of a given fit can be very well approximated by a quadratic fit, greatly cutting down the time that would be required for an exhaustive search.\\
  Using this quadratic, we can estimate the optimal weight for this pair of models, and combine the predictions linearly using this weight.
  \item Next, we compute correlations between the current ensemble and remaining models, and once again select the model with the lowest correlation.
  \item We return to step 3, using the ensemble and candidate model as models A and B.
  \item The loop is repeated until all candidate models have been added to the ensemble, and final used weights are calculated.
\end{enumerate}
As the algorithm adds models iteratively, it can be terminated early to gain almost all the total score with a subset of the models.
\newline

\subsection{Mixture of Experts for Ensembling DNNs}
\label{sec:moe}
In this section, we describe the approach of using mixture of experts for ensembling different DNNs. There are numerous methods commonly used for learning MoEs, and we refer the reader to \cite{Masoudnia2014} for a complete review. For our purpose, we have chosen to ensemble the outputs of different pre-trained DNNs using a gating network, whose weights are learnt using stochastic gradient descent.
%

We consider the following configurations of mixture of experts: Single coefficient per DNN; class-dependent linear combination coefficients and a dual-stream model that combines both approaches.

\subsubsection{Single Coefficient MoE}
\label{sec:sing_coeff_moe}
It is possible to learn the ensembling coefficients by posing it as a single layer convolutional network (Fig. \ref{fig:ensembling-nets}a). The input to this network, $\mathbf{I}$,  is the concatenation of the outputs of $N$ base DNNs into an $1\times C \times N$ tensor. The ensembling can be achieved by introducing a convolutional layer of a single filter of size $1\times1$ (with depth $N$). The output of this layer would be a tensor of size $1 \times C$. As such, the convolutional process performs the linear combination, with the weights of the filter acting as the ensembling weights. These outputs can then be compared against the corresponding labels using the cross entropy loss. The convolutional filter weights were learnt using the Adam algorithm.

\subsubsection{Per-class Coefficient MoE}
\label{sec:perclass_moe}
It is possible to increase the modelling capacity of the ensembling method by introducing more coefficients. To this end, we can introduce separate coefficients for each class and DNN (Fig. \ref{fig:ensembling-nets}b). As before, this can re-defined as a DNN learning problem. In this case, it is not possible to use convolutional filters. Instead, we introduce a new layer where its $\mathbf{W}$ weight tensor is the same size as the input tensor ($1\times C \times N$). An element-wise multiplication between this $\mathbf{W}$ and the input is then performed and summing carried out across the last dimension to produce an output of $1 \times C$. The output can then be compared with the corresponding labels using the cross-entropy loss and the weights learnt using the Adam algorithm.

\subsection{Regularised Dual-Stream MoE}
\label{sec:dual_stream_moe}
One shortcoming of the per-class coefficient MoE is the possibility of overfitting. While a single coefficient per DNN with far fewer parameters runs little risk of overfitting, it instead suffers from potentially being too restrictive. 

As such, we also explore a novel dual-stream model that combines both approaches. In this model, we aim to have the ensembling coefficients for a particular DNN centered around some ``mean'' value (i.e. similar to the single coefficient model), but also to add some small ``residual'' from this mean to provide some per-class flexibility.

The architecture of this model can be seen in Fig. \ref{fig:ensembling-nets}c. It consists of two initial parallel streams, each receiving the input tensor. The first stream is the $1 \times 1$ convolutional layer, providing the main ensembling coefficient for each DNN. The second stream is the per-class weight layer, providing the residual weight away from the first stream's coefficients. The outputs of both vectors (size $1 \times C$) are summed together providing the ensembled output vector.

In order to restrict the residual weights, we perform $l_2$ regularisation on the weight tensor of the second stream. This restricts the distribution of these weights to be zero mean with a small standard deviation.
Similar to the previous two approaches, the weights of both streams are then learnt by optimising the cross entropy loss function using the Adam algorithm. 

%
%

\section{Ensembling Experimental Results}
In this section, we compare performances of DNN ensembles with coefficients learnt using the methods described in Section \ref{sec:ensmb}. We present the performances of different DNN ensembles separately. For each of these approaches, we describe results on our validation dataset and the Kaggle test dataset. 
We have found that it was not possible to train the ensemble DNNs on the training dataset, due to the fact that the base DNNs are (unavoidably) overfitted to this set. Thus, using this training dataset for ensembling provides an inaccurate estimate of the generalisation performance of each model for the ensembler. For example, the fully connected DNNs have a much higher score on the training data, causing the ensembling algorithms to incorrectly converge on those and ignore the remaining DNNs.

As a result, it was necessary to use the validation dataset for learning the ensembling coefficients. 
However, the use of the entire validation dataset for learning ensembling coefficients leaves us with no remaining data for evaluating its generalisation performance. One approach would be to use the entire validation set for learning ensemble weights which are then used to ensemble and upload the Kaggle test data to its website for evaluation. However, this would be too costly time-wise.

In order to address the above issue, we have chosen to split the validation dataset into two partitions, which we denote as ensemble-train/test splits. We will then learn the DNN ensemble coefficients on the ensemble-train and test on the ensemble-test partition. The ensembling coefficients will be learnt on the ensemble-train dataset and evaluated on the ensemble-test partition.
\begin{figure}[h]
\centering
\includegraphics[width=0.47\textwidth]{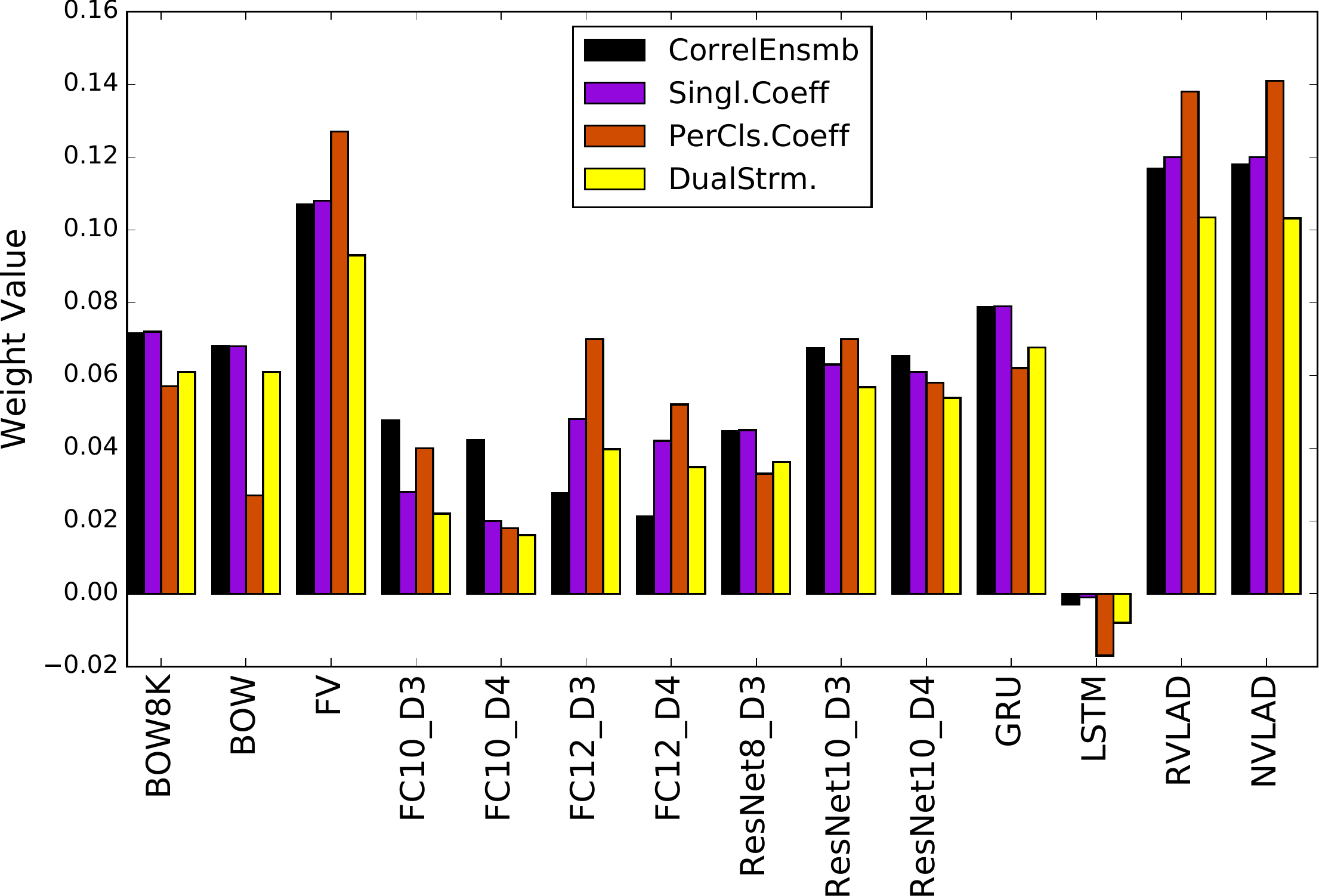}
\caption{DNN weights for the different ensembling methods (Sec. \ref{sec:ensmb}).
 }
\label{fig:all_ensmb_w}
\end{figure}

\subsection{Learnt Ensemble Weights}
We can see how the different ensembling methods have weighted the base DNN models in Fig. \ref{fig:all_ensmb_w}. Here, we observe that the LSTM model is given negative weighting. This is an artefact of the weight learning algorithms proposed here, and is caused by not constraining the weights to be non-negative. However, the negative weights assigned to the LSTM indicates that it does not contribute to improving the GAP20 score of the ensemble. This was confirmed by removing the LSTM model from the ensemble, and observing that the resulting performances on the Youtube8M dataset reported in Sections \ref{sec:dnn_generalisation} and \ref{sec:kaggle_test_res} were found to be very similar.

\begin{figure*}[h]
\centering
\begin{tabular}{ccc}
\includegraphics[width=0.35\textwidth]{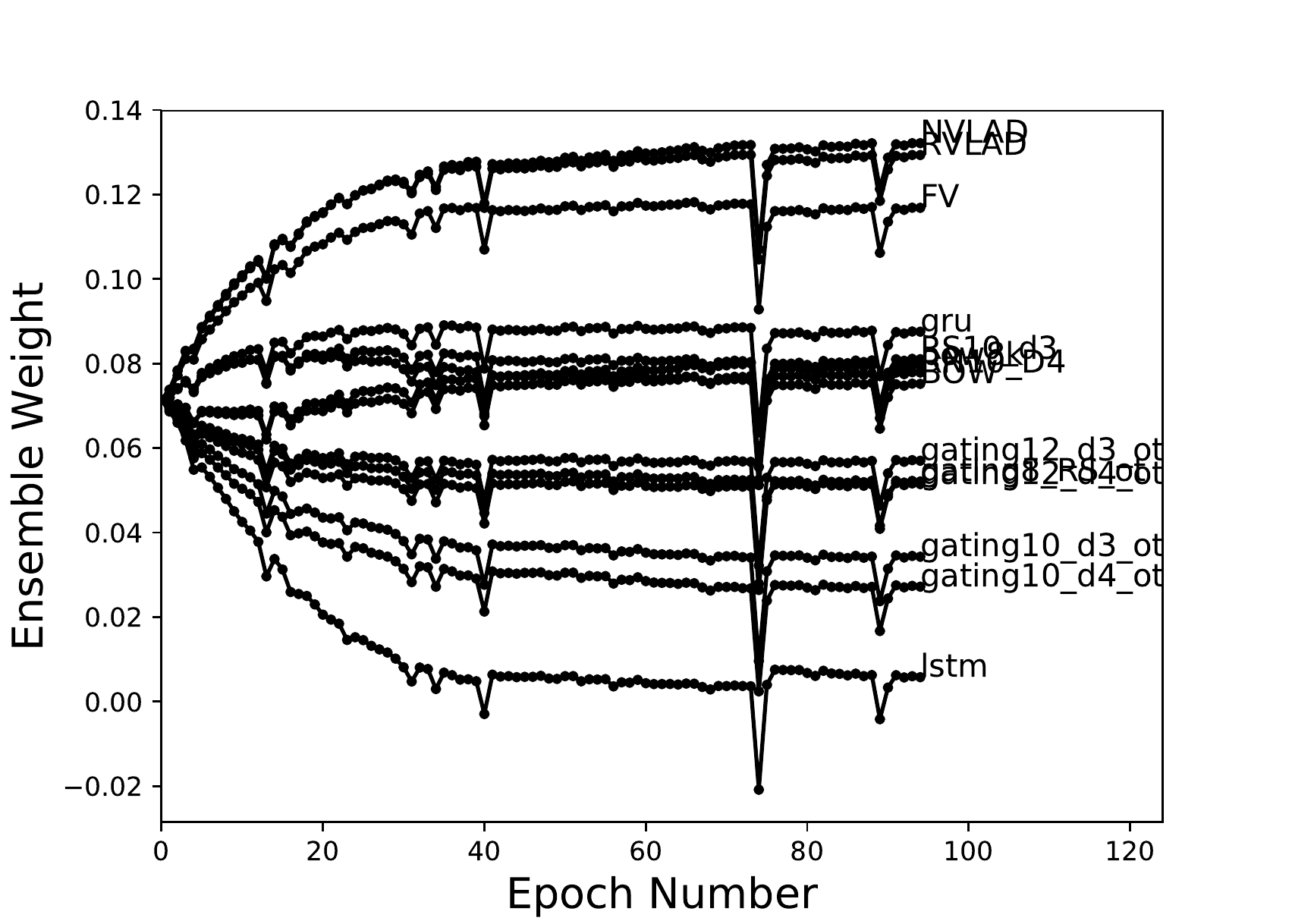} & 
\includegraphics[width=0.35\textwidth]{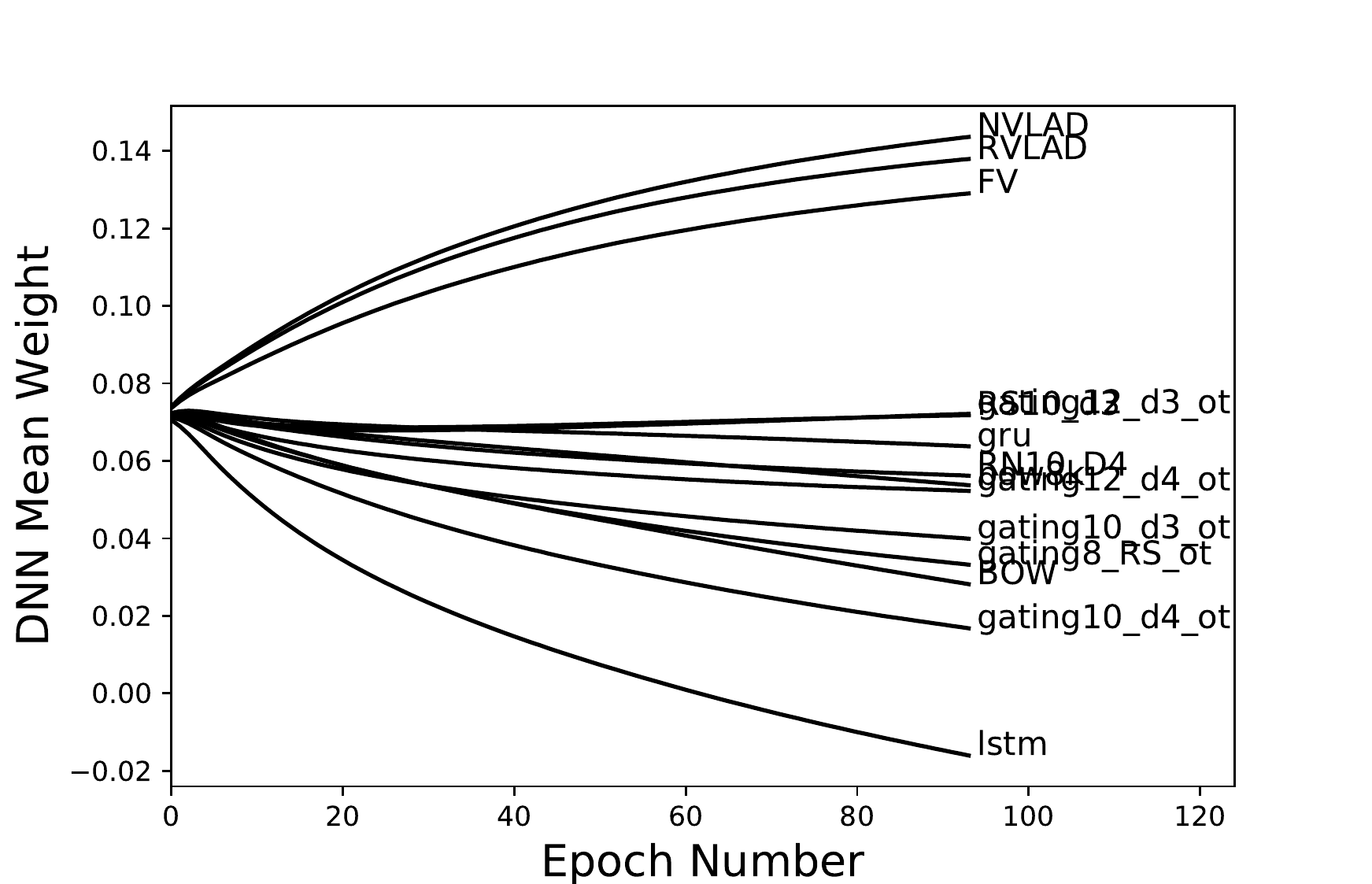} & 
\includegraphics[width=0.33\textwidth]{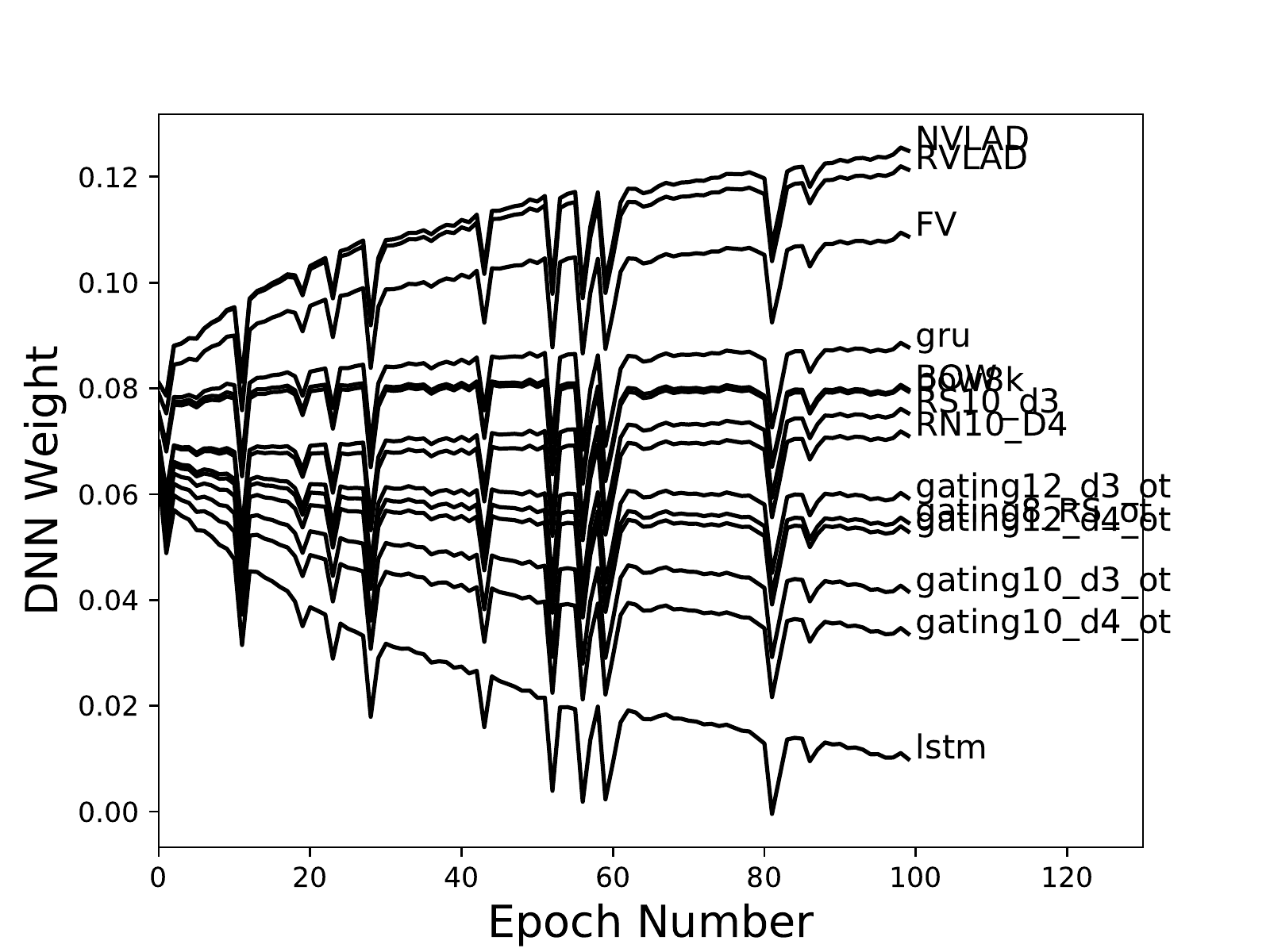} \\
 (a) Single Coeff. DNN & (b) Per-Clas Coeff. DNN & (c) Dual Stream DNN
\end{tabular}
\caption{ Evolution of DNN weights (or their means) across different epochs of the three different ensembling methods. }
\label{fig:ensmb_weight_evol}
\end{figure*}

For the per-class and dual-stream ensembling DNNs, we instead obtained the mean weights per individual base DNN. We see that all the methods have assigned approximately the same weights to different base DNNs. As expected, in general, the higher the individual performance of the DNNs, the higher the weights. Of particular interest is the negative weighting assigned by all the ensemblers to the LSTM method. We can see the evolution of the weights for each DNN in Fig. \ref{fig:ensmb_weight_evol}. For the single coefficient DNN emsembler, we find that the weights converge after approximately 40 epochs (Fig. \ref{fig:ensmb_weight_evol}a). An interesting artefact we have found is when the Adam algorithm performs large modifications to the weights, resulting in sudden jumps in weight values assigned to each DNN. However, this change is consistent across all DNNs.
For the per-class coefficient DNN ensembler, each DNN is not assigned a single weight, but instead 4716 separate weights, each for a class. However, we can see the trend of these weights by obtaining its mean value (for each DNN) at every epoch. This is shown in Fig. \ref{fig:ensmb_weight_evol}b. 

Finally, we show the weights assigned to the mean stream of the dual-stream DNN ensembler in Fig. \ref{fig:ensmb_weight_evol}c. For this method, the mean weights of the residual stream for each base-DNN is approximately 0 due to the L2 regularisation employed here.  The weights for each DNN evolve at a slower rate compared with the single coefficient DNN. However, we notice similar spikes in the weights due to the Adam algorithm.

\subsection{DNN-based Ensembling Generalisation}
\label{sec:dnn_generalisation}
\begin{figure*}[h]
\centering
\begin{tabular}{cc}
\includegraphics[width=0.35\textwidth]{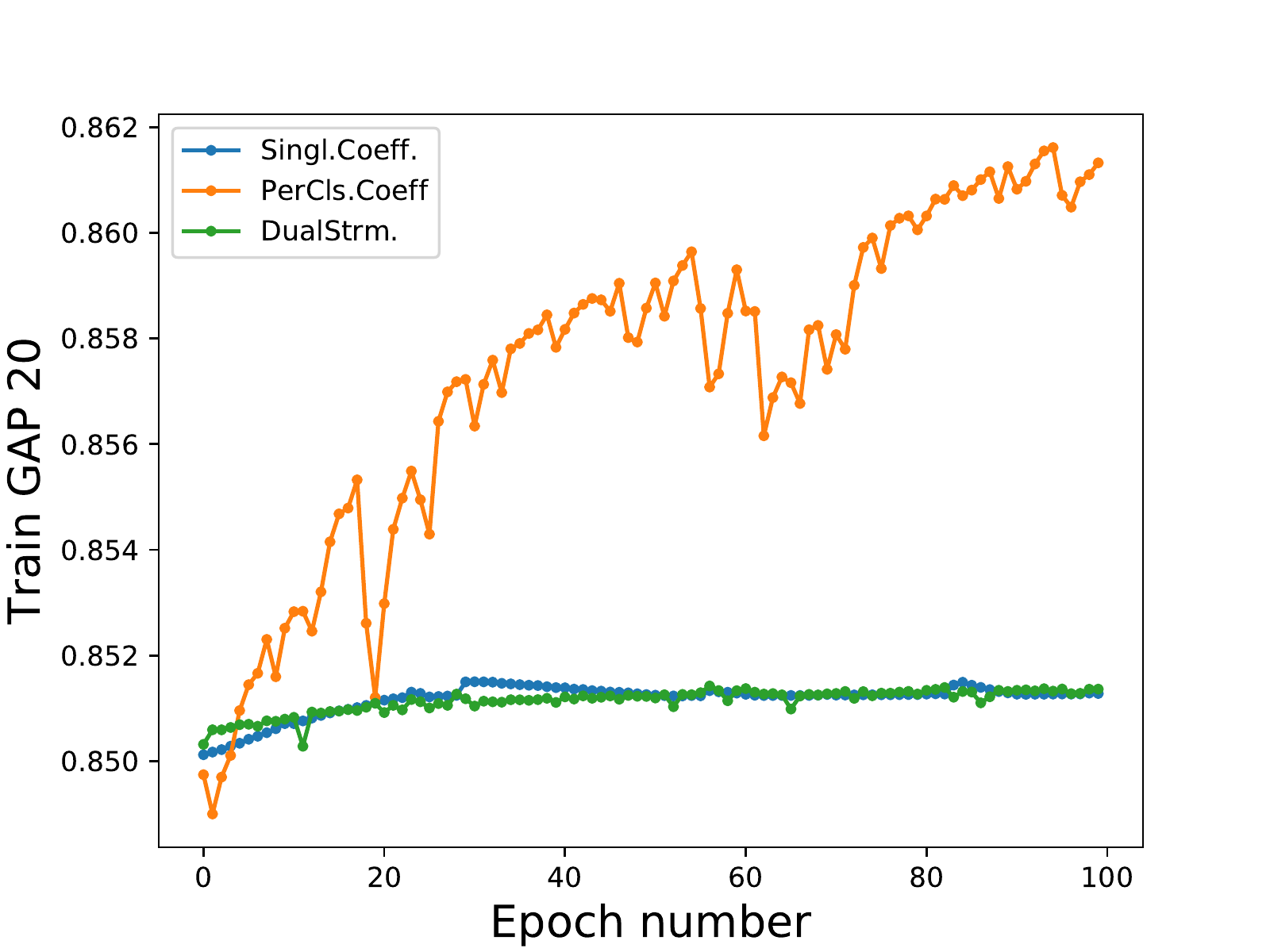} & 
\includegraphics[width=0.35\textwidth]{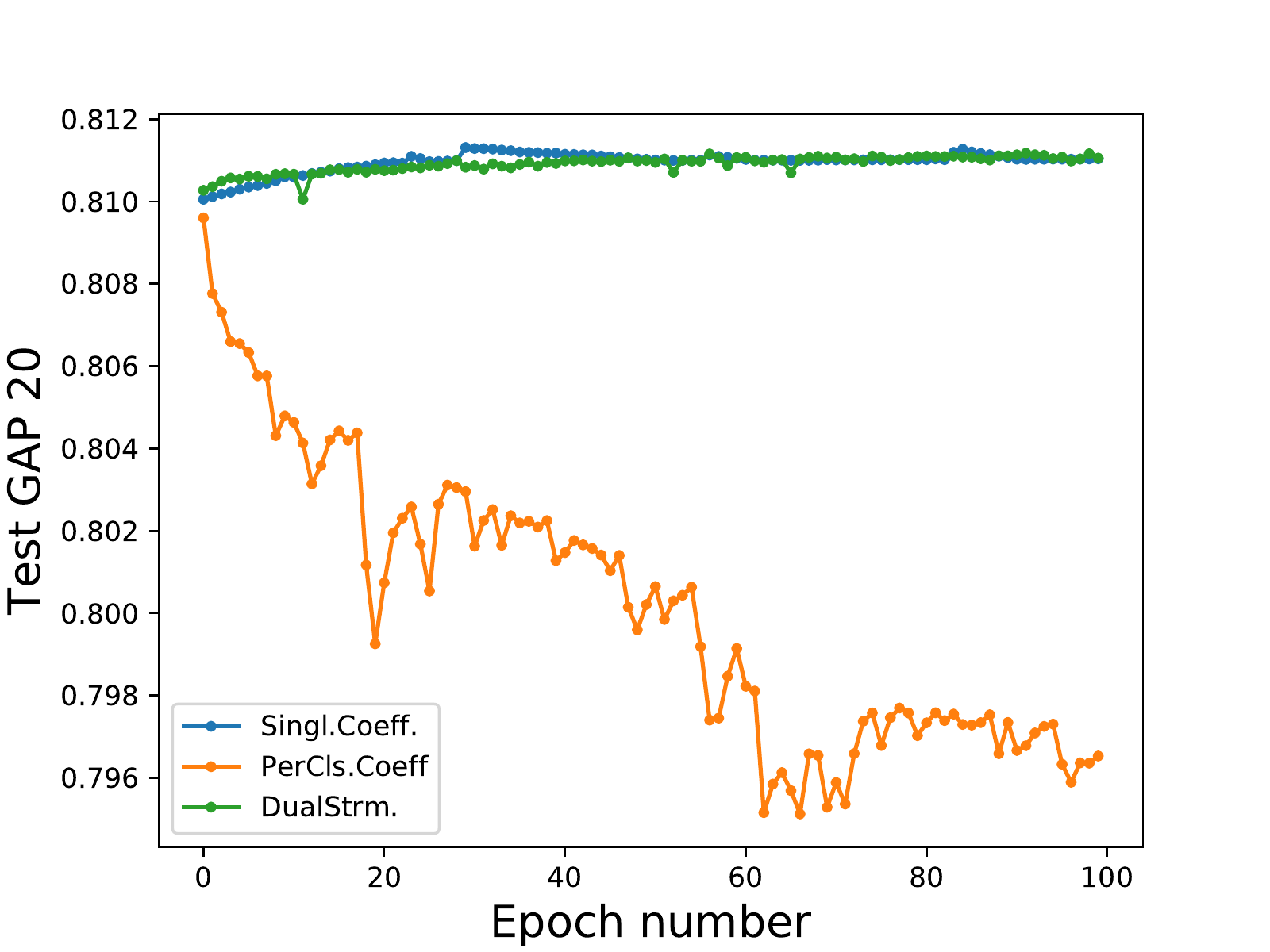} 
\\
(a) Training GAP20 & (b) Test GAP20
\end{tabular}
\caption{ The ensemble-train (a) and ensemble-test (b) GAP20 across all epochs for each of the proposed ensembling methods. }
\label{fig:train_gap_epoch}
\end{figure*}

In order to shed light on the generalisation capabilities of the different DNN-based ensembling approaches, we analyse the results from the training and test splits. These can be seen for all three ensembling methods in Fig. \ref{fig:train_gap_epoch}. It can be seen that the train and test results for both single coefficient and dual-stream approaches are roughly the same. 
However, we find that the per-class coefficient approach shows strong indications of overfitting. In particular, the training GAP20 score is seen to dramatically improve with increasing epochs. Unfortunately, we find that the test GAP20 score decreases in an equally dramatic fashion. 

\begin{figure*}[h]
\centering
\begin{tabular}{ccc}
\includegraphics[width=0.35\textwidth]{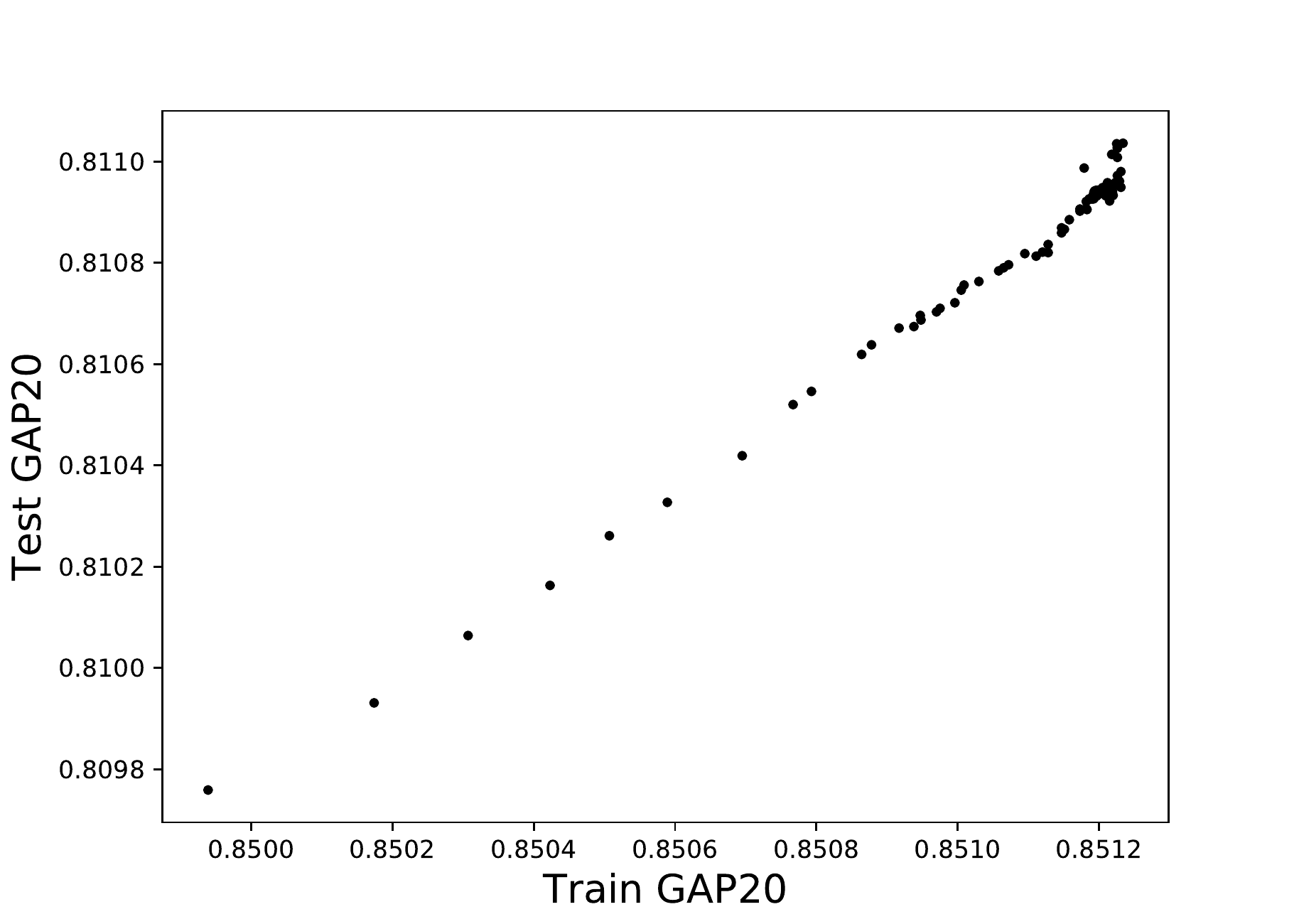} & 
\includegraphics[width=0.33\textwidth]{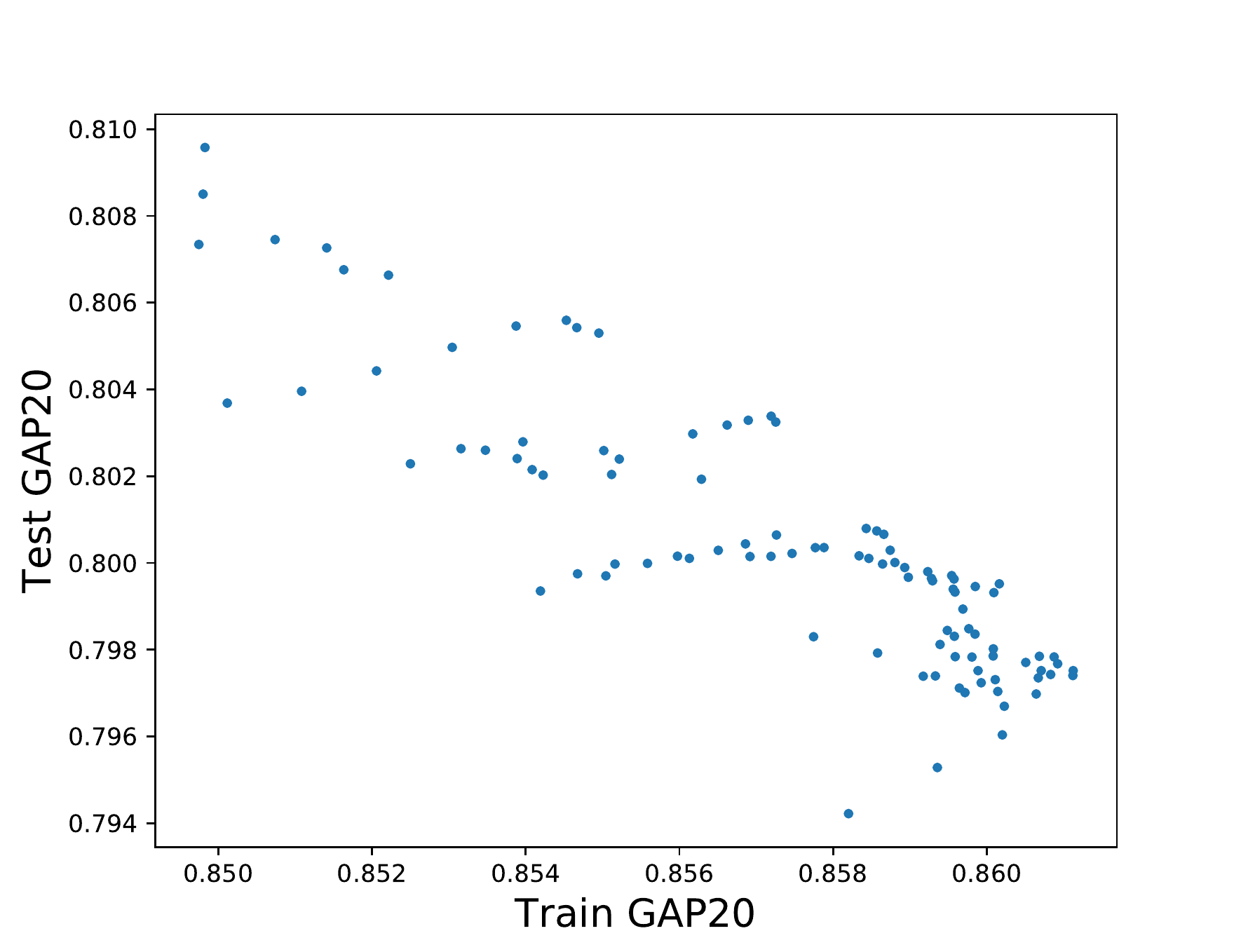} & 
\includegraphics[width=0.33\textwidth]{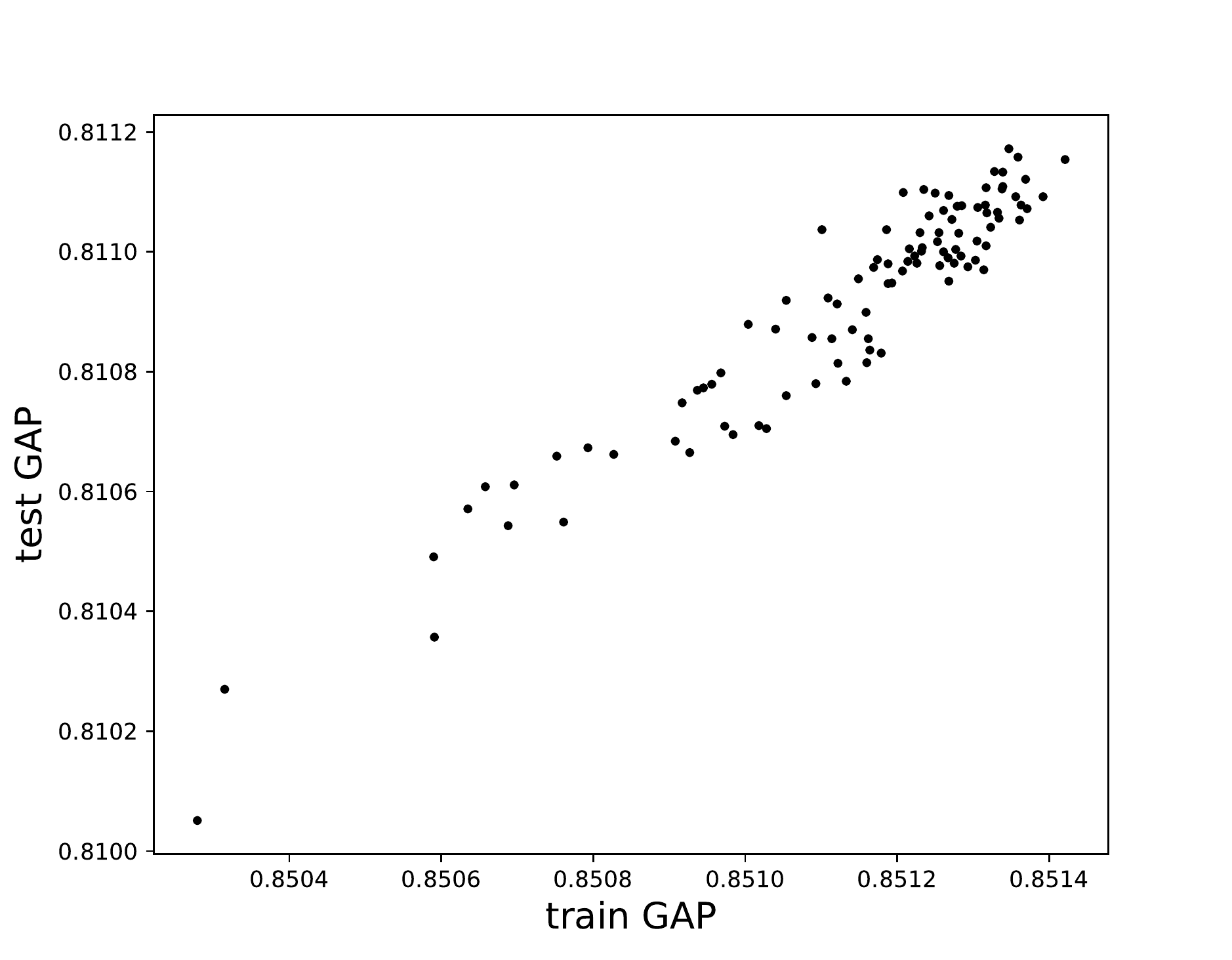} \\
 (a) Single Coeff. DNN & (b) Per-Clas Coeff. DNN & (c) Dual Stream DNN
\end{tabular}
\caption{ The scatterplots of training GAP20 vs test GAP20 for the three DNN-based ensembling methods.}
\label{fig:train_test_scatter}
\end{figure*}

This can be more clearly seen by producing the scatterplot of training vs test GAP20 scores across all epochs (Fig. \ref{fig:train_test_scatter}). 
In this figure, we see a strong correlation between increases in training and test GAP20 for the single coefficient ensembling DNN. In constrast, we find that there is an inverse correlation between the train and test GAP20 scores for the per-class ensembling method. Finally, for the dual-stream method, we still observe good correlation between train and test GAP increases. These suggest that the single coefficient method stands the highest chance of obtaining significant improvement in the GAP20 performance over the other methods.


\begin{table}[h]
    \centering
    \begin{tabular}{|c|c|c|c|}
    \hline
    Ensemble-Type     &  Ensemble Train GAP20 & Kaggle GAP20\\ \hline \hline
    Average & 85.03 & 84.97\\
    Correlation     & 85.12 & 85.11 \\
    Single Coeff. DNN & 85.13 &   85.11 \\
    Per-Class DNN & 86.16 &  83.83 \\
    Dual-Stream DNN & 85.12 &   85.10\\
    \hline
    \end{tabular}
    \caption{GAP20 scores on the ensemble training dataset and Kaggle-test dataset for different DNN ensembling methods.}
    \label{tab:ensmb_perf}
\end{table}

\subsection{Kaggle Test Results}
\label{sec:kaggle_test_res}
In order to produce the predictions for the Kaggle-test data, we chose to learn the ensembling coefficients using {\em all} the validation data. For this reason, we will denote the validation dataset as the {\em ensemble training dataset}. The resulting GAP20 scores on this dataset can be seen in Table \ref{tab:ensmb_perf}. 
We know that the per-class DNN score will be unreliable from the analysis of Section \ref{sec:dnn_generalisation}. However, we expect the other ensemble training dataset scores to closely reflect the performance gain compared to simple averaging of multiple DNNs in the unseen Kaggle-test data. The resulting GAP20 test results from the Kaggle website confirm this as can be seen in the third column in Table \ref{tab:ensmb_perf}. 

The baseline of the ensembling approach is obtained by simple averaging of all the outputs of the base DNNs. A baseline on the unseen Kaggle test dataset is obtained by averaging the base DNN results and uploading to the Kaggle website. The reported test GAP20 score obtained is 84.977\%. Both the correlation-based approach and single coefficient DNN ensembler achieved significant improvement of 85.11\% GAP20 results.
This overfitting issue of the per-class DNN is confirmed by a Kaggle test GAP20 score of 83.83\%. Finally, the dual stream method achieved a Kaggle-test GAP20 of 85.10\%. However, whilst this is a significant improvement compared with simple averaging, it does not improve on the single coefficient models.




The improvements obtained in the ensemble dataset can be verified by ensembling the unseen Kaggle test data using the learnt weights. This gives a Kaggle test GAP20 of 85.108\% over the score of 84.977\% when averaging is used. We have found that the Kaggle test GAP20 of 85.108\% could also be achieved by removing LSTM from the DNN ensemble. This is not suprising as the ensemble weight given to LSTMs was minimal (Fig. \ref{fig:all_ensmb_w}).

\section{Analysis of DNN-Ensemble Performance}
\label{sec:transfer_exp}
In the previous section, we have shown that ensembling provides significant improvement on the overall GAP20, both on the validation datset and the Kaggle test GAP20. In this section, we analyse the improvements brought about by the ensembling. To this end, we employ the same GAP20-based class accuracy scores described in Section \ref{sec:class_dep_dnn_perf}. We can then extract how accurate a particular model (base DNN or ensemble) at recommending some class label for videos. Similar to the analysis on individual DNNs, we consider the top 100 classes.

\begin{figure*}[h]
\centering
\begin{tabular}{cc}
\includegraphics[width=0.47\textwidth]{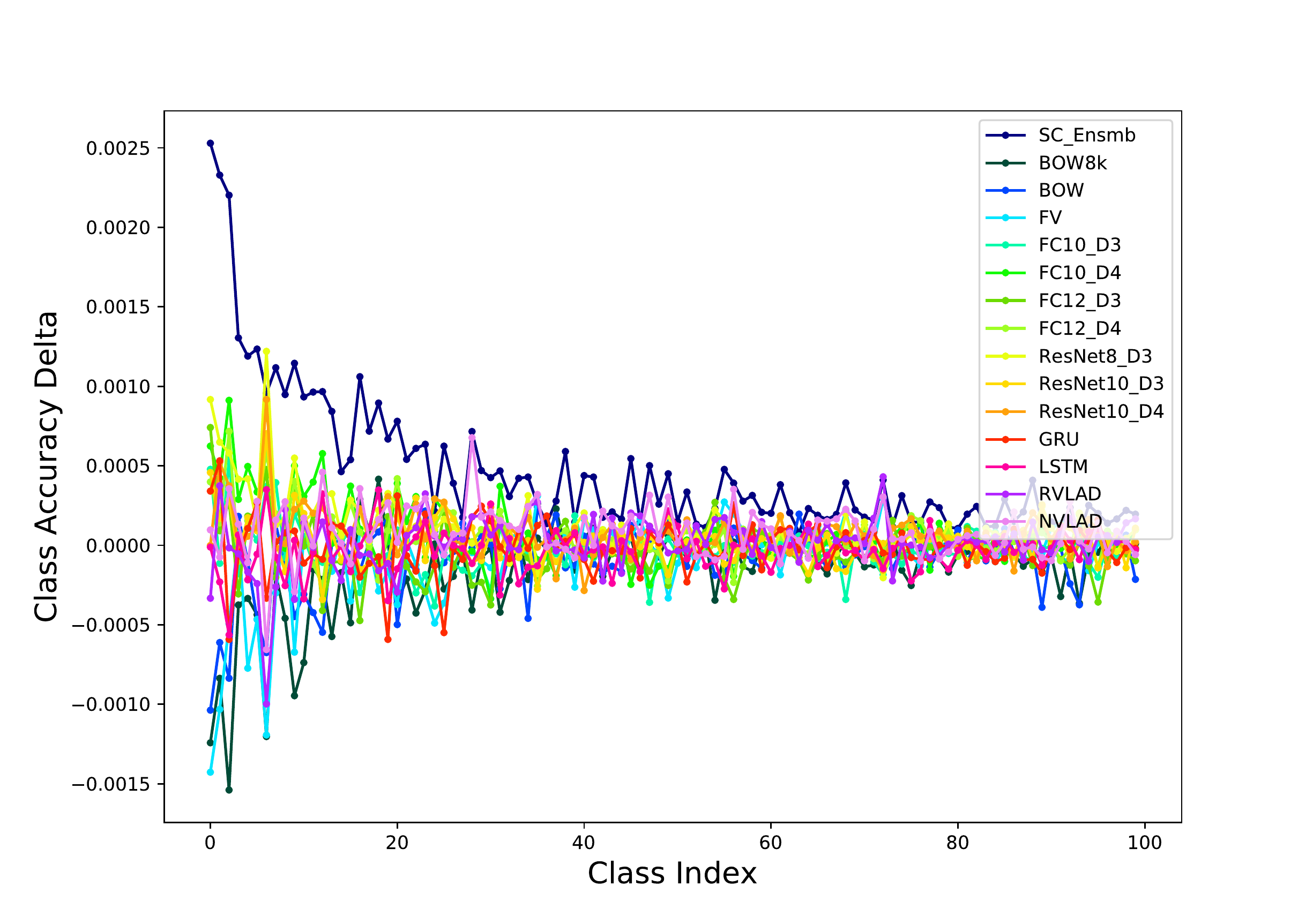} &
\includegraphics[width=0.47\textwidth]{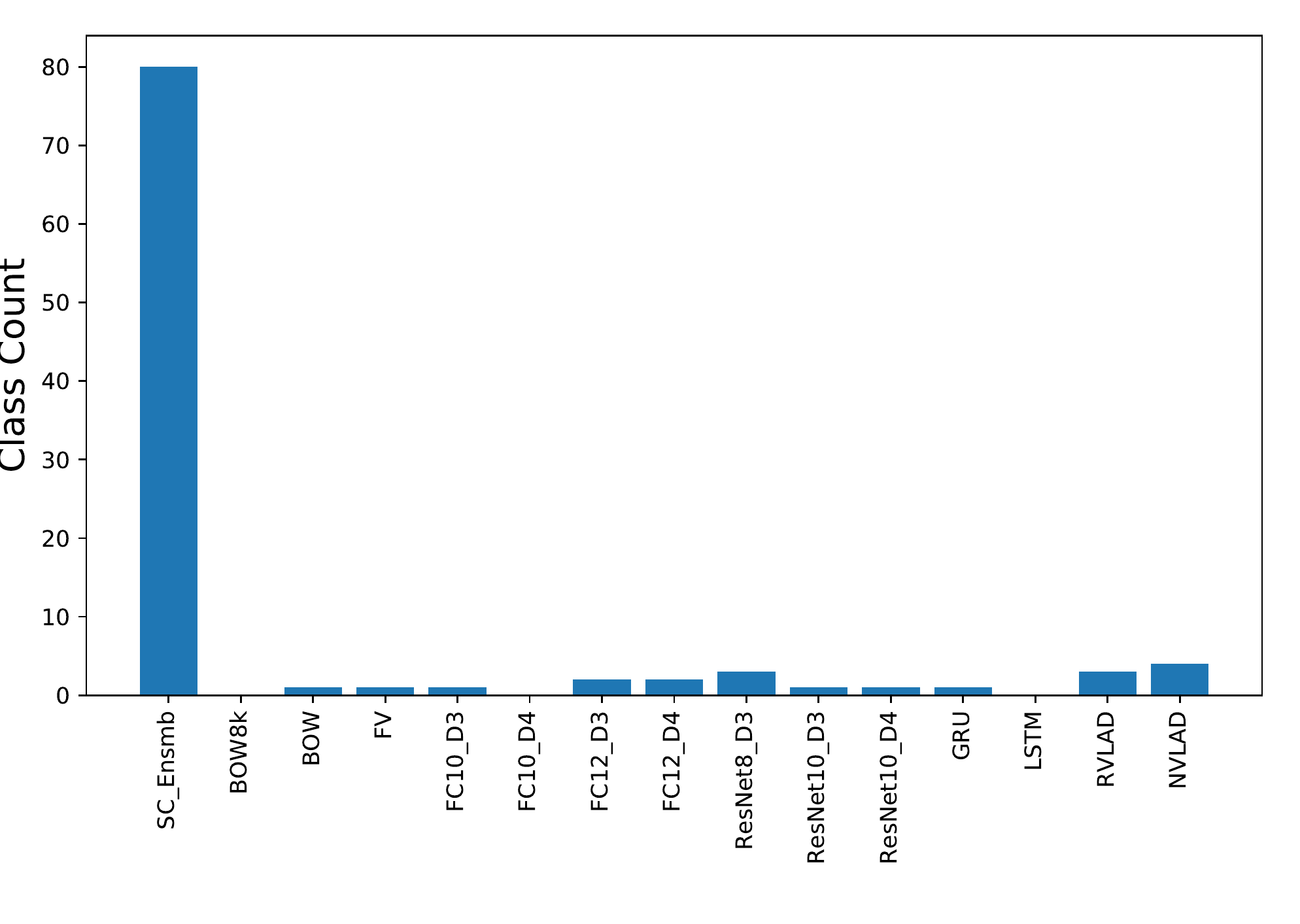} \\
(a) & (b)
\end{tabular}
\caption{ (a) This graph shows the difference between classifier accuracy and the mean accuracy across all base DNNs. (b) This graph shows that in the top 100 classes, the ensemble is the most accurate for approximately 80\% of the classes.
}
\label{fig:ensmb_analysis}
\end{figure*}

As before, to more clearly see the performance differences between different classifiers, here between base DNNs and the ensemble, we shall plot their difference to the mean accuracy of the base DNNs. This can be seen in Fig. \ref{fig:ensmb_analysis}a. We find that the single coefficient DNN ensemble is significantly more accurate than individual DNNs for the most frequent classes. In fact, the more frequent the class is, the greater the accuracy improvement.

Following this, we count of the number of classes the ensemble achieves improved accuracy when compared with all other classifiers. The results and can be seen in Fig. \ref{fig:ensmb_analysis}b. As expected, we find that the ensemble is the classifier that has maximum accuracy for the majority (79) of the top-100 classes.

\section{Transfer Learning on Related Datasets: UCF101 and HMDB51}
In order to allow for comparison with existing work, we have performed transfer learning on the DNNs used for the Youtube8M dataset on two benchmark datasets: UCF101 \cite{ucf101} and HMDB51 \cite{hmdb}. Both datasets consists of a number of videos from various public databases such as YouTube and Google. The UCF101 dataset consists of 101 action categories with 13320 videos. The HMDB51 dataset contains 51 action categories, with 7000 videos. Additionally, for cross validation testing purposes, both datasets have three train-test partitions.

For both the UCF101 and HMDB51 datasets, we have selected the best DNNs from the Youtube8M dataset for use in classification: gated VLAD, gated RVLAD, gated netFV, gated BOW, Gated FC 10K, Gated ResNet 8K and Gated ResNet 10K. 

In order to perform transfer learning, all the layers following the first gating layer were removed and initialised from random. The remaining layers prior to the gating layer were initialised with the weights trained using the Youtube8M dataset. The temporal agnostic DNNs were fine-tuned for 500 epochs, whilst the fully connected DNNs were fine-tuned for 20 epochs.

In order to learn the coefficients for ensembling, a separate set of DNNs were trained using the training data with 10\% reserved as validation data. The single coefficient ensembling method and validation data was used for obtaining the ensembling coefficients. Following this, the coefficients were used for ensembling the DNNs that were trained on all the training data. This ensured we were using the maximum amount of training data available for generating the final predictions.

The ensembled system had an accuracy of 84\% on the UCF101 dataset. Detailed results on the individual DNN performances can be seen in Table \ref{tab:ucfhmdb}. Here, we find that the ResNet architectures perform better than the other methods. We can see that an improvement of approximately 2\% over the best individual DNN performance was obtained by ensembling.
The performance of 84\% for the UCF101 dataset is comparable to existing state-of-the art methods when RGB full-frame features are used: 83.35\%(MVSV)\cite{MVSV}, 84.78\%(Conv. Layer Pooling,VGGNet)\cite{convpool}, 86.0\%(Hyp-Net,RGB).

In terms of the HMDB51 dataset, we have found that the ensemble of DNNs had an class accuracy of 52.6\%. The detailed performances of different DNNs is also shown in Table \ref{tab:ucfhmdb}. As with the UCF101 dataset, the ResNet archictures again obtains the highest accuracies. We have found that ensembling different DNNs resulted in an increase of approximately 1.8\% in accuracy. Similar to UCF101, performance of the proposed ensemble is similar to state-of-the-art methods:
55.9\%(MVSV)\cite{MVSV}, 50.42\%(Conv. Layer Pooling,VGGNet)\cite{convpool}, 54.8\%(Seq.VLAD)\cite{seqvlad}.

Interestingly, we find that for both the UCF101 and HMDB51 datasets, using a compact representation of 1152-D feature vectors for each 1-second duration of visual information in a video can result in classification accuracies comparable to when the entire RGB image of every video frame is used. Even more interesting is that, good performance can be obtained when only a {\em single} feature vector (2304-D) (mean and standard deviation of the above 1152-D feature vectors) is used to represent an entire video.

\begin{table}[h]
    \centering
    \begin{tabular}{|c|c|c|}
    \hline 
    Method & UCF101 & HMDB51 \\
    \hline 
    Gated FC 10K & 80.89\% & 48.11\% \\
    Gated ResNet 8K & 82.18\% & 59.26\%\\
    Gated ResNet 10K & 82.10\% & 50.85\% \\
    \hline
    Gated netBOW & 58.74\% & 35.88\% \\
    Gated netVLAD & 79.80\% & 37.91\%\\
    Gated netRVLAD & 80.47\% & 43.20\% \\
    gated netFV & 80.94\% & 40.20\%\\
    \hline
    Ensemble & 84.06\% & 52.61\% \\
    \hline
    \end{tabular}
    \caption{Class Accuracy for different DNNs and ensemble on the UCF101 and HMDB51 datasets}
    \label{tab:ucfhmdb}
\end{table}

\subsection{Analysis}
In both datasets, we have found that performing ensembling of different DNNs have resulted in significant improvements in classification accuracy. 
Additionally, we find that the performance of the ensemble is similar to related methods where optical flow was not used. Importantly, the proposed method only requires 1152-dimensional features as opposed to all RGB frames of the entire video clip. This allows the methods and their ensemble used in this paper to scale well to extremely large datasets such as Youtube8M, both in terms of computational and storage requirements. 

We have also found that the networks based on the mean and standard deviation features outperformed the temporal agnostic DNNs. One reason for this is potentially due to the short time durations of the UCF101 and HMDB51 clips, averaging around 5-10 seconds. In comparison, the Youtube8M videos are considerably longer in length, averaging at about 229 seconds \cite{45619}. Since the temporal agnostic models are cluster-based and trained on the much longer Youtube8M dataset, the number of clusters used will likely be too many for a clips that are 1/50 the length. Consequently, many clusters will be redundant, adversely affecting the fine-tuning process. We observe this in the reduced accuracy of these models when compared with the fully connected NN architectures. These DNNs use the mean feature vectors, which tend to be less sensitive to differences in video clip lengths, thus allowing better performance.

\section{Conclusions}
\label{sec:conclusions}
In this paper, we have evaluated the performance of a wide range of different DNN architectures and video features. The different architectures ranged from using video-level statistics to frame-level recurrent networks such as LSTMs. Additionally, we have also proposed a novel DNN architecture based on ResNet that achieves state of the art performance for individual classifiers, whilst using substantially simpler video-level statistical information. We have found that, on an individual level, each DNN achieves roughly the same range of GAP20 accuracy, from 82\% to 83\%. Nonetheless, analysis on their individual performances across different classes to show that they are diverse. Additionally, two ensemble level measures of diversity provide further indication that adding more classifiers will increase diversity. This in turn suggests a strong case for ensembling the different classifiers. 

We proposed four different methods for performing DNN ensembling. These approaches differ mainly by the number of linear combination weights used for the ensembling process. We have found that the ensembling method with only a single coefficient assigned to each base-DNN has the best generalisation capability. When more weights are used, with each class and base-DNN assigned a different weight, overfitting becomes a serious issue. Whether adding more training data will overcome this remains an open question. We have found that the single coefficient ensemble of 13 diverse base DNNs gives us the state-of-the-art GAP20 performance from the Kaggle website of 85.12\%. This is in contrast with the existing highest score of 84.9\% obtained by simple averaging of 25 DNNs models. Analysis of the performance of the ensemble indicates that there is significant improvement in labelling accuracy for the most frequent classes.

Additionally, we have performed transfer learning on existing DNNs from the above on the UCF101 and HMDB51 datasets, and demonstrated that our methods are similar in performance to state-of-the-art methods despite using features-vectors that are considerably more compact when compared with full RGB frames. Ensembling has also been demonstrated to result in significant improvements on these datasets.


%

{\small
\bibliographystyle{ieee}
\bibliography{egbib}
}








\end{document}